\documentclass[11pt]{article}

\usepackage[preprint]{acl}

\usepackage{times}
\usepackage{latexsym}

\usepackage[T1]{fontenc}
\usepackage[utf8]{inputenc}

\usepackage{microtype}

\usepackage{inconsolata}

\usepackage{graphicx}

\usepackage{tabularx}
\usepackage{multirow}
\usepackage{multicol}
\usepackage{pifont}
\usepackage{booktabs}
\usepackage[table]{xcolor}
\usepackage{algorithm}
\usepackage{algpseudocode}
\usepackage{amssymb}
\usepackage{placeins}
\usepackage{enumitem}
\usepackage{amsmath,amsfonts,bm}
\usepackage{cuted}

\graphicspath{{figures/}}

\newcommand{\cmark}{\textcolor{green!55!black}{\ding{51}}}
\newcommand{\xmark}{\textcolor{red!70!black}{\ding{55}}}  

\definecolor{mygreen}{RGB}{0, 176, 80}
\definecolor{myblue}{RGB}{68, 114, 196}

\newcommand{\hlprimarytab}[1]{\colorbox{yellow!20}{#1}}
\newcommand{\uadown}{\raisebox{0.5\depth}{\tiny$\downarrow$}}
\newcommand{\uaup}{\raisebox{0.5\depth}{\tiny$\uparrow$}}
\newcommand{\uagdown}[1]{{\scriptsize\hlprimarytab{\uadown{#1}\%}}}
\newcommand{\uagup}[1]{{\scriptsize\hlprimarytab{\uaup{#1}\%}}}

\usepackage{listings}
\usepackage{xcolor}

\lstdefinestyle{mystyle}{
    language=Python,
    basicstyle=\ttfamily\scriptsize, % 使用更小的等宽字体以适应窄栏
    breakatwhitespace=false,         
    breaklines=true,                 % 开启自动换行
    captionpos=b,                    
    keepspaces=true,                 
    showspaces=false,                
    showstringspaces=false,
    showtabs=false,                  
    tabsize=2,                       % 缩进设为2个空格，节省横向空间
    frame=single,                    %哪怕换行了也包裹在框里
    keywordstyle=\color{blue},       % 关键词高亮（可选）
    stringstyle=\color{purple},      % 字符串高亮（可选）
    commentstyle=\color{gray},
    postbreak=\mbox{\textcolor{red}{$\hookrightarrow$}\space}, % 换行处显示一个小箭头提示
}

\title{From Inertia to Objectivity: Improving Deep Research \\Agents with Noise Isolation}

\author{
  \textbf{Xiangxin Zhang\textsuperscript{1,2}},
  \textbf{Zhanwei Zhang\textsuperscript{3}},
  \textbf{Zhihang Fu\textsuperscript{4}},
  \textbf{Binbin Lin\textsuperscript{1}},
  \textbf{Wenxiao Wang\textsuperscript{1,2}}\thanks{Corresponding author.}
\\
\\
  \textsuperscript{1}School of Software Technology, Zhejiang University \qquad
  \textsuperscript{2}Darwinmind AI \\
  \textsuperscript{3}State Key Lab of CAD\&CG, Zhejiang University \qquad
  \textsuperscript{4}Alibaba Group \\
  \texttt{\{xiangxinzhang, zhanweizhang, binbinlin, wenxiaowang\}@zju.edu.cn} \\
  \texttt{zhihang.fzh@alibaba-inc.com}
}

\begin{document}
\maketitle
\begin{abstract}
Web search agents powered by Large Language Models (LLMs) show strong promise, but deep research tasks expose a recurring failure mode: once an agent has produced a query, plan, or intermediate conclusion, it becomes less objective when later judging the consequences of that same action. We term this phenomenon \textbf{inertia bias}. To make it measurable, we introduce the IBIS benchmark, which controls the search observations while varying whether the model is evaluating the outcome of its own prior action. We find that models are substantially worse when they ``own'' the preceding search step, showing that self-authored action history can systematically distort subsequent judgment. We further show that this bias propagates into two forms of system-level degradation: search noise at the worker level and contextual noise at the manager level. To address this problem, we propose NIS-Agent, which applies context isolation at the two decision points most vulnerable to inertia bias: webpage triage and final-answer validation. Across GAIA, WebWalkerQA, BrowseComp, and BrowseComp-zh, NIS-Agent achieves competitive performance while reducing token cost by 33\% compared to our baseline. We further train an 8B model to be intrinsically more resistant to inertia bias; under the same NIS-Agent framework, it attains average performance comparable to GPT-4o on deep research benchmarks. Our code is publicly available at \url{https://github.com/PangSMPang/NIS-Agent}.
\end{abstract}

\section{Introduction}

Large Language Models (LLMs) have rapidly evolved from simple predictors into powerful autonomous agents capable of multi-step reasoning, tool use, and complex planning~\citep{java2025characterizingdeepresearchbenchmark,guo2024largelanguagemodelbased}. These agent frameworks often employ iterative planning and execution loops~\citep{yao2023react,Significant_Gravitas_AutoGPT}, and PlanBench frameworks~\citep{NEURIPS2023_7a92bcde}, augmented by memory mechanisms for long-horizon tasks~\citep{packer2024memgptllmsoperatingsystems,xu2025amemagenticmemoryllm}. For existing web search agent frameworks, manager--worker communication patterns~\citep{zhang2024chainagentslargelanguage} are commonly adopted. The manager devises high-level strategies and decomposes tasks, while the worker gathers relevant information for each subtask. Both the manager and the workers operate under the \textit{thought--action--observation} paradigm (abbreviated as \textbf{T}, \textbf{A}, and \textbf{O}, respectively). Specifically, the manager formulates a plan (T), then invokes its own tools or the workers (A), and after receiving feedback (O), it continues iterating until it deems that a final answer has been obtained. For the worker, it first generates a query (T), then calls a tool to search for relevant webpages (A), obtains search results (O), selects the most relevant page (T), browses it (A), and acquires information (O), repeating this loop until enough evidence is collected to report back to the manager.

\begin{figure*}[t]
\begin{center}
\includegraphics[width=\textwidth]{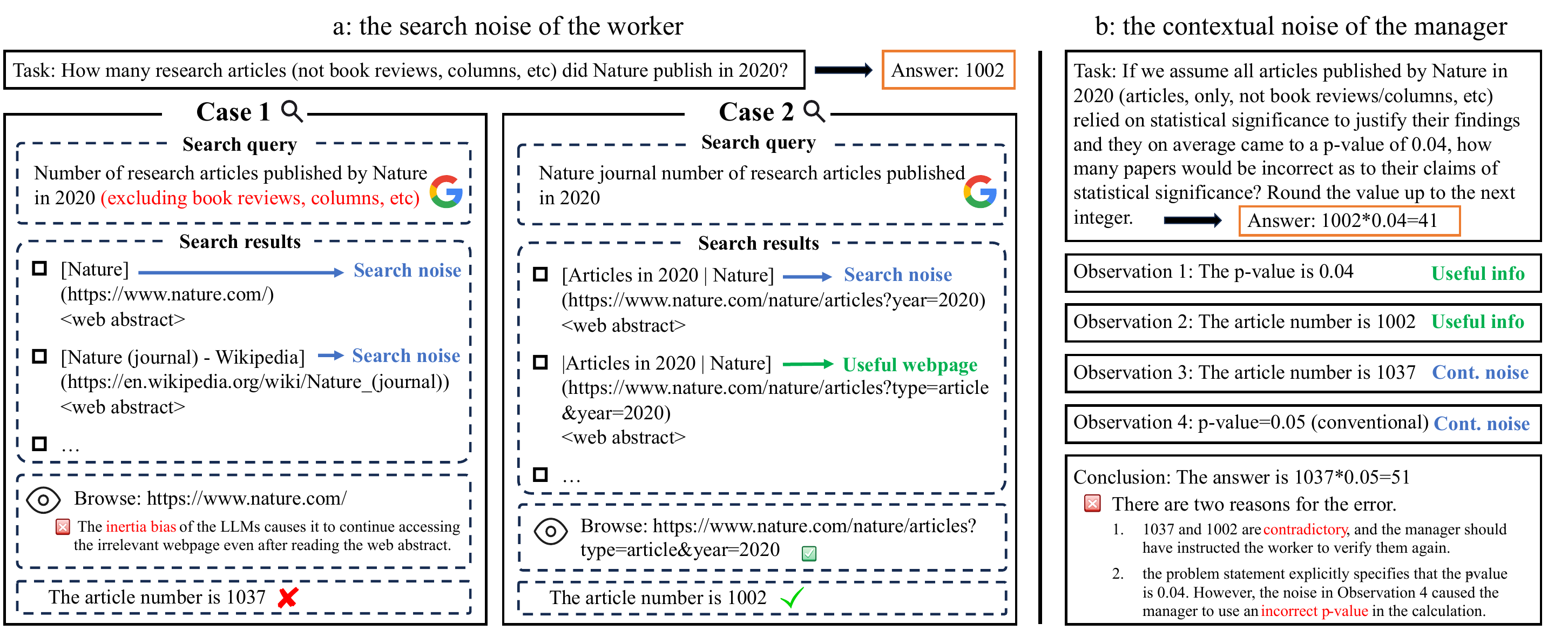}
\end{center}
\caption{Two kinds of noise. Words in \textcolor{mygreen}{green} and \textcolor{myblue}{blue} indicate useful information and noise, respectively. The words in \textcolor{red}{red} reveal the causes of the two types of noise. In subfigure (a), the \texttt{<web abstract>} denotes the summary information of webpages returned by Google search. For two seemingly similar queries, the results differ drastically. In Case~1, due to the inherent inertia bias of the LLMs, the worker continues searching along the query, and arrives at an incorrect answer. In subfigure (b), we omit the manager’s thought and action steps, retaining only the observation results. The contextual noise misleads the manager into faulty reasoning, ultimately yielding an incorrect answer.}
\label{fig:twonoise}
\end{figure*}

While these frameworks have achieved significant success, substantial limitations persist when they operate within open network environments, especially for deep research (see Section~\ref{deepresearch}) tasks that require maintaining extensive contextual states. In this setting, we identify a specific agentic failure mechanism: LLMs tend to reinforce their own prior choices regardless of whether those choices were correct. We refer to this phenomenon as \textbf{inertia bias}. Concretely, once an agent has produced a query, plan, or intermediate conclusion, it becomes less objective when later judging the consequences of that same action. To rigorously investigate and quantify this phenomenon, we introduce \textbf{IBIS (Inertia Bias in Information Seeking)}, a diagnostic benchmark designed to isolate the impact of contextual history on decision-making. IBIS specifically targets the information-seeking workflow of the worker, which involves generating a query, executing a search, and deciding whether to browse a page or revise the query. We construct scenarios where a query returns irrelevant search abstracts, making a re-search the only factually correct action. We then evaluate models under two controlled conditions. In Agentic Mode, the context includes the model's own history of issuing that query. In Observer Mode, the same search results are presented as external information. We find that models in Agentic Mode fail significantly more often than in Observer Mode, demonstrating that LLMs lose objectivity when evaluating their own action history.

This quantified inertia bias amplifies \textbf{noise} at two critical decision points in deep research agents. First, it generates \textbf{search noise} at the worker level (Figure~\ref{fig:twonoise}(a)). Open network environments contain an overwhelming amount of heterogeneous information, and web search is highly sensitive to the wording of a query. After the SearchAgent steps onto an unproductive path, it could in principle reject that path in time by examining webpage abstracts alone. In practice, inertia bias makes the agent continue browsing pages that appear superficially relevant but are actually unhelpful. Second, the same self-reinforcing tendency creates \textbf{contextual noise} at the manager level (Figure~\ref{fig:twonoise}(b)). If the manager forms an inaccurate plan or interpretation in the early stage, it tends to preserve that view as context grows longer and noisier. Instead of objectively reassessing the accumulated evidence, it selectively ``cherry-picks'' observations that appear compatible with its earlier thought, which can lead to premature termination or incorrect final answers.

To address this failure mode, we propose the Noise Isolation Search Agent (NIS-Agent), a web search framework designed to \textbf{mitigate noise accumulation by restoring objectivity at key decision points}. The main idea is not merely to reduce context length, but to \textbf{isolate judgments from the agent's own action history} whenever that history is likely to bias subsequent decisions. Concretely, we introduce two context-isolation mechanisms. (i) For SearchAgent, we propose a context-isolated filter module that evaluates candidate webpages using only the task-relevant state, enabling the agent to reject irrelevant search results before browsing and thereby reducing search noise. (ii) For ManagerAgent, we propose an isolation-based stepwise validation module that reorganizes the reasoning chain and validates local inferences under isolated context, thereby reducing contextual noise near the final decision stage.

We build our agent on top of smolagents \citep{smolagents}, and use the framework developed by the smolagents team for Deep Research, which we refer to as smolagents DR \citep{smolagents_open_deep_research}, as our initial baseline. Our experimental results demonstrate that NIS-Agent achieves a significant performance enhancement over the smolagents DR while reducing token cost by 33\%.  With stronger model configurations, NIS-Agent achieves state-of-the-art performance on both GAIA~\citep{mialon2023gaiabenchmarkgeneralai} and WebWalkerQA~\citep{wu2025webwalkerbenchmarkingllmsweb} benchmarks among open-source frameworks. Beyond mitigating inertia bias at inference time, we further train an 8B open-source model via SFT and GRPO to be intrinsically more resistant to it, yielding a small model whose deep-research performance is comparable to GPT-4o.

The contributions of our work are as follows:
\begin{itemize}
    \item We identify \textbf{inertia bias}, an agentic failure mechanism in which LLMs lose objectivity toward their own action history, and we construct the IBIS benchmark to quantify it.
    \item We propose NIS-Agent, which uses \textbf{context isolation} to mitigate the noise accumulation exacerbated by inertia bias.
    \item We further train an 8B open-source model to be \textbf{intrinsically} less susceptible to inertia bias, achieving deep-research performance competitive with closed-source large models.
\end{itemize}

\section{Related Work}

\subsection{Reasoning Bias and Self-Correction} 
\label{subsec:reasoning bias}

LLMs exhibit persistent cognitive biases that hinder objective reasoning. Sycophancy describes the tendency to align with \emph{external} user inputs rather than objective facts~\citep{perez2022discoveringlanguagemodelbehaviors, sharma2025understandingsycophancylanguagemodels}, while confirmation bias causes models to favor evidence that supports a prior belief, skewing rationale generation~\citep{wan2025unveilingconfirmationbiaschainofthought}. Inertia bias is distinct from both: its trigger is \emph{internal} and \emph{action-oriented}---the model becomes anchored to its own previously generated action (a query, plan, or intermediate conclusion) rather than to external human framing or a propositional belief. In agentic workflows, this broader family of biases undermines self-correction, as models often fail to detect errors within their own generated context~\citep{huang2024largelanguagemodelsselfcorrect}. While frameworks such as Reflexion~\citep{shinn2023reflexionlanguageagentsverbal} and CRITIC~\citep{gou2024criticlargelanguagemodels} introduce feedback loops to mitigate this, they typically operate within an accumulating context. We fundamentally decouple judgment from the model's \emph{own} action history and quantify the resulting loss of objectivity.

\subsection{Deep Research}
\label{deepresearch}
Deep research refers to conducting multi-step searches on the internet for complex tasks~\citep{openai2025introducing}. For existing deep research agent frameworks, manager--worker communication patterns~\citep{zhang2024chainagentslargelanguage} are commonly adopted. For the manager, reducing hallucinations and noise is crucial for improving system capability. For example, KnowAgent~\citep{zhu-etal-2025-knowagent} introduces an action-knowledge base to constrain action path generation, Agent KB~\citep{tang2025agentkbleveragingcrossdomain} utilizes a teacher model to guide the student model in searching and providing suggestions, and Agent Workflow Memory~\citep{wang2024agentworkflowmemory} reuses summarized sub-workflows. However, all of these methods rely on external experience databases.  
For the worker, ASCoT~\citep{zhang2025ascotadaptiveselfcorrectionchainofthought} and Slow-Thinking~\citep{gan2025rethinkingexternalslowthinkingsnowball} intervene at critical steps, while WebShaper~\citep{tao2025webshaperagenticallydatasynthesizing} employs compositional Knowledge Projection (KP) operations to finely control the reasoning structure. Their common objective is to detect issues as early as possible and correct the action trajectory.
Although these directions also modify how context is used, their main emphasis is memory reuse, action guidance, or early correction within the standard agent trajectory. By contrast, our work focuses on a distinct question: how to recover objectivity once the agent has become biased by its \emph{own} action history. Accordingly, we use context isolation not as generic compression, but as a targeted debiasing intervention.

\begin{figure*}[t]
\begin{center}
\includegraphics[width=\textwidth]{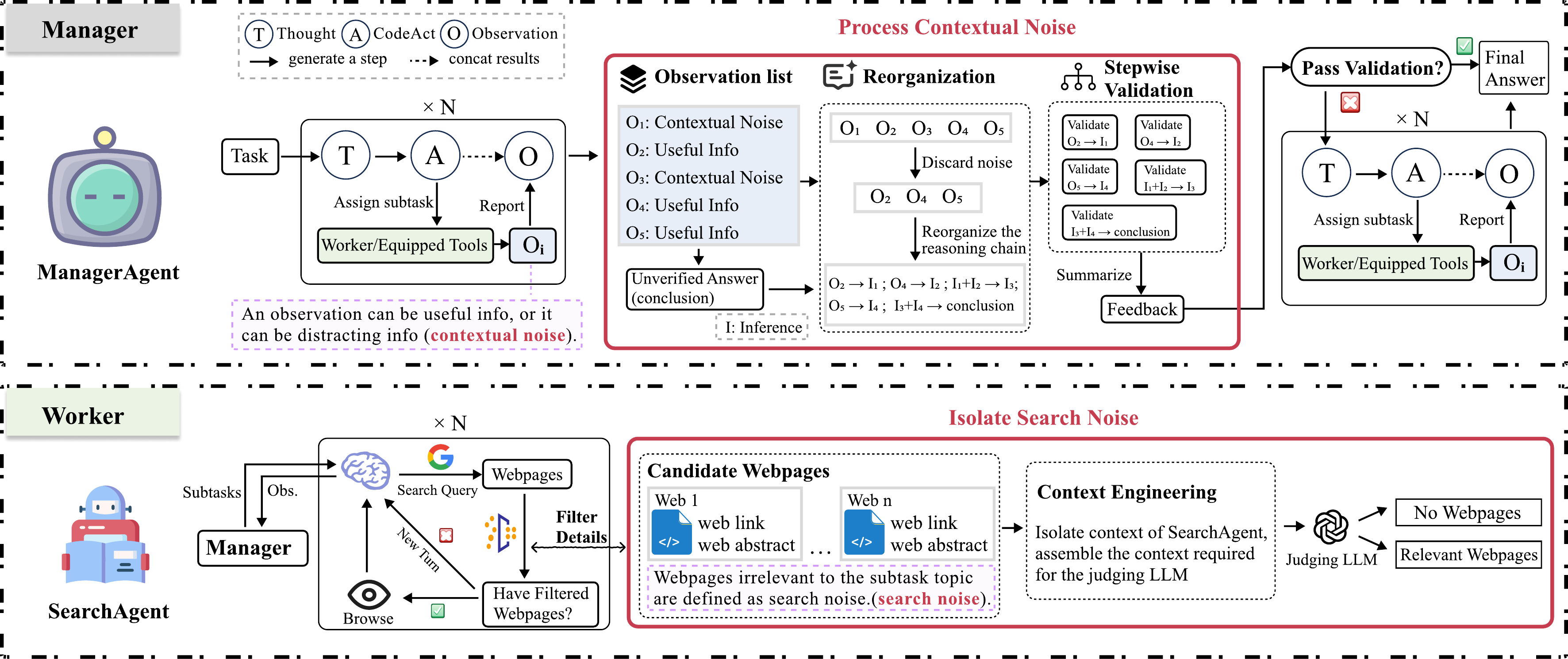}
\end{center}
\caption{Framework of NIS-Agent. The two components highlighted in \textcolor{red}{red} in the figure constitute the main focus of our optimization. \textbf{Manager}: ManagerAgent alternates between thought, codeact, and observation. The actions include invoking Worker or calling equipped tools. It continuously accumulates contextual noise during iteration. When it deems the answer ready, the system automatically triggers the validation module (see Section~\ref{sec:validation}). If the answer passes validation, it is output immediately; otherwise, the ManagerAgent resumes the iteration. \textbf{Worker}: SearchAgent is responsible for collecting information from the web. We specifically optimized the filter stage (see Section~\ref{sec:filter}) to mitigate search noise. Once collecting enough information, SearchAgent will organize and report it.}
\label{framework}
\end{figure*}

\section{Inertia Bias in Information Seeking}
\label{sec:ibis}

To rigorously investigate and quantify inertia bias in LLM-based agents, we introduce the \textbf{IBIS (Inertia Bias in Information Seeking)} benchmark, a diagnostic benchmark designed to isolate the impact of action history on decision-making.

\subsection{Inertia Bias as an Agentic Failure Mechanism}

In our search pipeline, the ideal behavior is straightforward: the agent generates a query, inspects the returned abstracts, and abandons the current search path when those results are clearly irrelevant. However, when the agent evaluates those results together with its own preceding action history, the actual behavior differs systematically from this ideal. The model tends to select URLs even though the summaries already indicate that the current query is unproductive. This behavior reflects a form of path dependence to the agent's own prior action.

As discussed in Section~\ref{subsec:reasoning bias}, this phenomenon is conceptually distinct from both sycophancy and confirmation bias. The source of influence is not external user framing, as in sycophancy, nor merely commitment to a prior belief, as in confirmation bias. Instead, the model becomes anchored to its own previously generated query, plan, or intermediate conclusion. For deep research agents, this distinction matters because the same self-reinforcing mechanism affects both page triage and answer validation. It therefore creates a unified failure mode that manifests as search noise at the worker level and contextual noise at the manager level.

\subsection{Dataset Construction}
We first prompt qwen3-max~\citep{qwen3max} to generate 1,016 candidate factual question tasks across multiple domains. These tasks require specific, verifiable answers, such as statistics, dates, or measurements. For each task theme, the model also generates search queries with varying levels of specificity to simulate real-world user queries. Three independent LLMs, namely Gemini-2.5-Pro, DeepSeek-R1, and Claude-3.7-Sonnet, then score each candidate along three dimensions: the factual verifiability of the question, the uniqueness of the answer, and the consistency between the query and the information need. This scoring step yields 712 high-quality samples.

For each retained sample, we execute its query against Google Search. Three annotators then independently categorize the results, based on the retrieved web abstracts, into one of three classes. The first class is Should Re-search, where the abstracts are clearly irrelevant to the information need. The second class is Should Visit Page, where at least one result appears relevant based on its title and snippet. The third class is Should Return Answer, where the abstract already contains the answer. We keep only the samples on which all three annotators agree, and we discard the rest. This way, annotation quality and consistency are controlled without an additional arbitration step. The resulting agreement, measured by Fleiss' $\kappa$, is 0.684, indicating substantial agreement. IBIS targets the decision point most directly related to inertia bias, namely whether to continue or abandon the current search path. We therefore retain only the first two categories, yielding 245 Should Re-search samples and 209 Should Visit Page samples. Full details of this protocol are provided in Appendix~\ref{appendix:annotation_protocol}.

\subsection{Evaluation Protocol}
The IBIS benchmark is established to evaluate two dimensions of agentic performance. First, to assess general reasoning capabilities in determining the optimal next step, we utilize the Full Set, measuring the model's baseline accuracy in distinguishing whether search results are sufficient (warranting a page visit) or insufficient (necessitating a re-search). Second, to quantify the specific magnitude of inertia bias, we isolate the \textit{Should Re-search} subset. In this subset, sticking to the current search path is factually incorrect. Therefore, a failure to reject the results serves as a direct indicator of the model's irrational adherence to its previous action.

To rigorously decouple the influence of action history from content reasoning, we evaluate models under two controlled conditions with identical system prompts:
\textbf{Agentic Mode}: The context includes a conversational history where the model itself explicitly calls the \texttt{web\_search} tool. The search results are presented as the direct observation of this self-initiated action, forcing the model to evaluate the outcome of its own decision.
\textbf{Observer Mode}: The identical task and search results are provided as external reference information without an assistant turn indicating self-initiation. This places the model in a neutral stance, detaching the search results from its own authorship.

\begin{figure*}[t]
\begin{center}
\includegraphics[width=\textwidth]{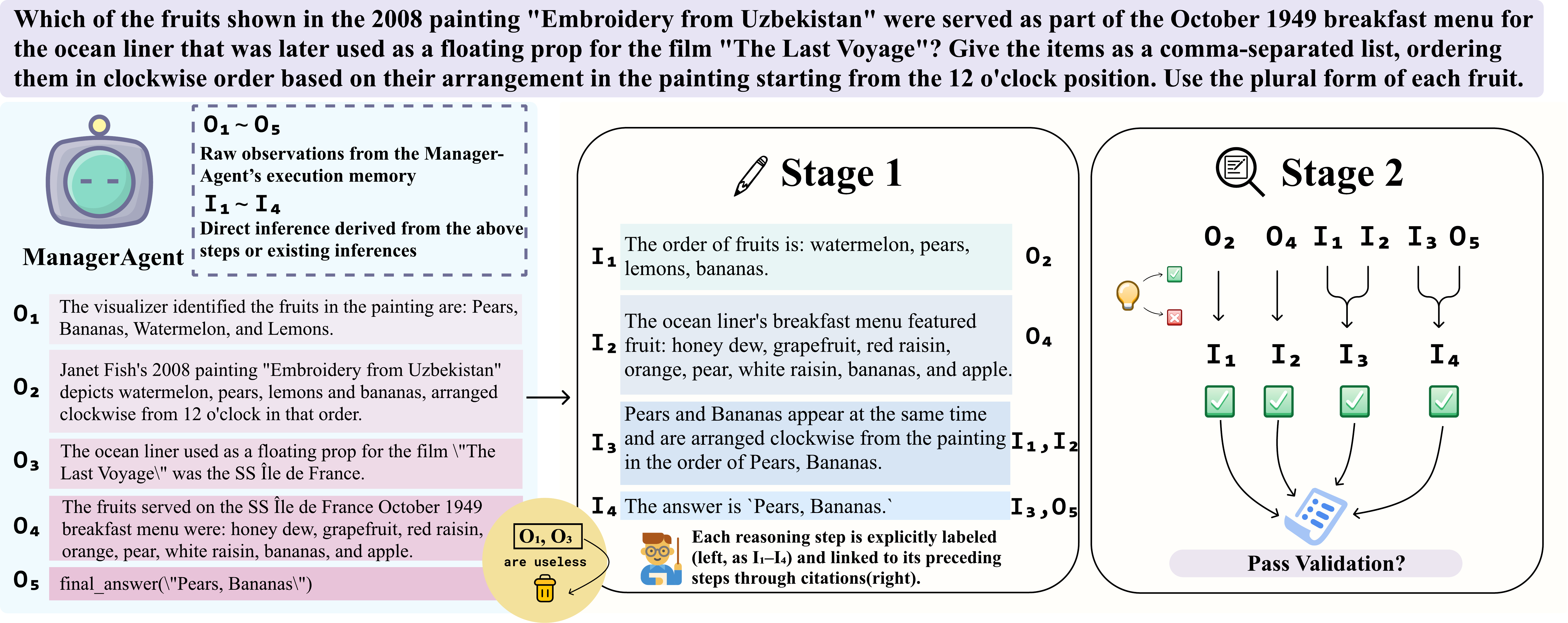}
\end{center}
\caption{Workflow of the isolation-based stepwise validation module. This module is automatically triggered when the ManagerAgent generates its final answer for the first time. Stage~1: The module reorganizes the reasoning chain behind the ManagerAgent's answer by extracting the raw execution history and pruning irrelevant context. In this process, each reasoning step is explicitly labeled, providing structured input for the next stage. Stage~2: For each step in the reasoning chain, we independently validate only the local inference between its premise and conclusion. For example, in $a \to b \to c$, we validate $a \to b$ and $b \to c$ separately. After obtaining all suggestions, they are fed back to the ManagerAgent as guidance for its subsequent actions (continue iterating or outputting the answer).}
\label{isolated_validation}
\end{figure*}

\section{NIS-Agent}
\label{introduce:nis-agent}
We propose the Noise Isolation Search Agent (NIS-Agent), a multi-agent framework for open-web deep research. Rather than redesigning the standard agent architecture, NIS-Agent intervenes at the two decision points most vulnerable to inertia bias: a context-isolated filter targets worker-level search noise during webpage triage, and an isolation-based stepwise validation module targets manager-level contextual noise at final-answer validation. Figure~\ref{framework} illustrates the framework.

\subsection{Context-Isolated Filter Module}
\label{sec:filter}
The SearchAgent issues a \texttt{web\_search} query and then chooses which returned webpages to browse. As established in Section~\ref{sec:ibis}, inertia bias makes this triage step unreliable: the agent tends to commit to webpages produced by its own query even when the abstracts already indicate they are irrelevant, wasting computation and injecting noise downstream.

To address this, we design a context-isolated filter module that screens candidate search results before webpage access. The key point is that the relevance judgment is deliberately separated from the SearchAgent's full execution trajectory. A standard in-context filtering step is still asked to reason inside the same history that produced the current query, and thus remains vulnerable to rationalizing that query. Our module instead reconstructs a compact decision context from the task-relevant state and evaluates candidate webpages under isolation. This allows the model to judge whether the current search path should be continued or abandoned with greater objectivity. When relevance is insufficient, the module triggers adaptive query rewriting, thereby improving retrieval efficiency and accuracy. Figure~\ref{fig:web_filter_main} illustrates the workflow of this module.

\subsection{Isolation-based Stepwise Validation Module}
\label{sec:validation}

Existing methods either directly output the answer~\citep{smolagents} or hand the entire reasoning trace to an LLM for validation~\citep{wu2025webwalkerbenchmarkingllmsweb}. Both reuse the same context that produced the answer, leaving the validation step exposed to inertia bias. We therefore introduce an isolation-based stepwise validation module, automatically triggered when the ManagerAgent first deems an answer ready. It audits the proposed solution in two stages. The workflow is shown in Figure~\ref{isolated_validation}.
\textbf{Stage 1: Reasoning Process Reorganization.}
In the first stage, the module extracts key reasoning steps from the ManagerAgent’s execution memory, including the original task description and the relevant tool-call sequences. It then reorganizes these scattered execution steps into a coherent reasoning chain. Each reasoning step is explicitly labeled and linked to its preceding steps through citations, ensuring that intermediate steps are properly referenced.
\textbf{Stage 2: Stepwise Reasoning Validation.}
In the second stage, the module independently verifies each reorganized reasoning step. To do so, it examines whether the inference logically follows from the referenced conditions by extracting these conditions, constructing \textit{condition-inference} pairs, and evaluating correctness step by step. For any problematic steps, the module provides concrete suggestions for improvement. If all steps are valid, it confirms the final answer as correct.
If the answer passes validation, the ManagerAgent produces the final answer. Otherwise, the ManagerAgent re-evaluates the provided suggestions with full context and decides whether to overrule them or proceed with another round of iteration.

\section{NIS-8B}
\label{sec:nis_8b}

NIS-Agent mitigates inertia bias at inference time on top of frontier closed-source models. We further ask whether a much smaller open-source model can be trained to be intrinsically less susceptible to inertia bias. We explore this question with a two-stage training pipeline on Qwen3-8B~\citep{qwen3technicalreport}, resulting in a model we name NIS-8B.

\subsection{Supervised Fine-tuning.}
The SFT stage targets two complementary capabilities. The first is procedural competence: producing well-formed tool calls and adhering to the structured output protocol expected by an agent loop. For this, we draw on a publicly available tool-use corpus~\citep{liu2024apigenautomatedpipelinegenerating}. The second is task-distribution warm-up for the subsequent RL stage: we synthesize decision examples following the IBIS construction protocol using a strong external teacher, supervising the model directly on gold actions. Since both data sources are disjoint from our evaluation benchmarks, the SFT stage does not contaminate the test sets we report on.

\subsection{Reinforcement Learning with Group-Relative Policy Optimization.}
SFT establishes an initial prior by imitating gold actions, but imitation alone cannot correct the model when it makes inertia-biased decisions during its own rollouts. We therefore continue training with GRPO~\citep{shao2024deepseekmathpushinglimitsmathematical} on top of the SFT checkpoint, allowing the model to internalize anti-inertia behavior through self-exploration and feedback.

Each training prompt places the agent immediately after its own first \texttt{web\_search} observation, so a rollout consists of a single tool-call decision: revise the query, browse a candidate page, or terminate. We define the reward for this decision as
$$R = 0.1 \cdot R_{\text{format}} + 0.9 \cdot R_{\text{action}},$$
where $R_{\text{format}}$ indicates whether the rollout adheres to the expected agent output protocol, and $R_{\text{action}}$ indicates whether the chosen action is judged correct against a pre-computed reference rubric (see Appendix~\ref{sec:nis8b_training}). Concentrating the reward at the exact decision point where inertia bias manifests provides a denser credit-assignment signal than evaluating the full trajectory's final answer. The 0.1 / 0.9 weighting follows the practice established by WebSailor~\citep{li2025websailornavigatingsuperhumanreasoning}. Training details are reported in Appendix~\ref{sec:nis8b_training}.

\section{Experiment}
\label{experiment}

\begin{table*}[t]
\centering
\caption{Main experimental results. The best results are highlighted in \textbf{bold}, and the second-best are \underline{underlined}. Results in \textcolor{gray}{gray} are our own reproductions, as they were not officially reported. Except for methods marked with $\ddagger$, where the \textit{Pass@n} metric was not explicitly stated in their official publications, all results are under the \textit{Pass@1} metric. $\dagger$ denotes that the method is evaluated on the text-only subset of the GAIA validation set (103 samples).}
\label{tab:main_results}
\scalebox{0.95}{
\renewcommand{\arraystretch}{1.1}
\begin{tabular}{llccclc}
\toprule
\multirow{2}{*}{\textbf{Method}} & \multirow{2}{*}{\textbf{Model}} & \multicolumn{4}{c}{\textbf{General AI Assistant (GAIA)}} & \textbf{WebWalkerQA} \\
\cmidrule(lr){3-6}
& & \textbf{Level 1} & \textbf{Level 2} & \textbf{Level 3} & \textbf{Avg.} & \textbf{Avg.} \\
\midrule
Agent-KB & GPT-4.1 & 79.25 & 58.14 & 34.62 & 61.21 & - \\
smolagents DR & GPT-4o & \textcolor{gray}{66.04} & \textcolor{gray}{55.29} & \textcolor{gray}{30.77} & \textcolor{gray}{54.88} & \textcolor{gray}{46.50} \\
smolagents DR & GPT-4.1 & \textcolor{gray}{69.81} & \textcolor{gray}{60.47} & \textcolor{gray}{34.62} & \textcolor{gray}{59.39} & \textcolor{gray}{53.00} \\
OWL & Claude-3.7-Sonnet & \textbf{84.91} & 68.60 & 42.31 & 69.70 & - \\
OAgents & Claude-3.7-Sonnet & 77.36 & 66.28 & 46.15 & 66.67 & - \\
WebExplorer\textsuperscript{$\ddagger$} & Claude-Sonnet-4 & - & - & - & 68.30\textsuperscript{$\dagger$} & 61.70 \\
BrowseMaster\textsuperscript{$\ddagger$} & DeepSeek-R1-0528 & - & - & - & 68.00\textsuperscript{$\dagger$} & 62.10 \\
AWorld & Claude-Sonnet-4 & - & - & - & 67.89\textsuperscript{$\dagger$} & - \\
MiroFlow & GPT-5 & - & - & - & 71.90 & 52.60 \\
MemoBrain & GLM-4.6 & 79.50 & 71.20 & \underline{50.00} & 71.80 & 66.50 \\
\rowcolor[rgb]{0.902,1,0.902} NIS-Agent(Ours) & GPT-4o & 81.13 & 58.14 & 34.62 & 61.82 & 57.50 \\
\rowcolor[rgb]{0.902,1,0.902} NIS-Agent(Ours) & GPT-4.1 & 83.02 & 67.44 & 38.46 & 67.88 & 59.00 \\
\rowcolor[rgb]{0.902,1,0.902} NIS-Agent(Ours) & Claude-3.7-Sonnet & \underline{84.62} & \underline{72.41} & \underline{50.00} & \underline{72.73} & \underline{68.50} \\
\rowcolor[rgb]{0.902,1,0.902} NIS-Agent(Ours) & DeepSeek-V4-Pro & \textbf{84.91} & \textbf{83.72} & \textbf{65.38} & \textbf{81.21} & \textbf{75.00} \\
\rowcolor[rgb]{0.902,1,0.902} NIS-Agent(Ours) & NIS-8B & 83.92 & 63.95 & 26.92 & 63.64 & 59.00 \\
\bottomrule
\end{tabular}
}
\end{table*}

\subsection{Experimental Setup}
\label{sec:exp_setup}

\paragraph{Benchmarks.}

We evaluate our method on two primary deep-research benchmarks: (i) GAIA~\citep{mialon2023gaiabenchmarkgeneralai}, a widely-adopted benchmark for General AI Assistants, with tasks spanning multimodal analysis, tool use, web search, and complex reasoning. This benchmark is organized into three difficulty levels from 1 (easiest) to 3 (hardest). (ii) WebWalkerQA~\citep{wu2025webwalkerbenchmarkingllmsweb}, which evaluates the ability of agents to browse subpages. In addition, to assess broader transferability, we further report results on BrowseComp~\citep{wei2025browsecompsimplechallengingbenchmark} and BrowseComp-zh~\citep{zhou2025browsecompzhbenchmarkingwebbrowsing}. Finally, to show that inertia bias is not specific to deep research, we also study AIME 2024~\cite{aime2024} and AIME 2025~\cite{aime2025} in a separate reasoning-only setting.

For GAIA, we use its entire public validation set, which consists of 165 queries. For WebWalkerQA, BrowseComp and BrowseComp-zh, we randomly sample 200 examples, consistent with the settings used by most baselines. We adhere to the experimental setup of Webthinker~\citep{li2025webthinkerempoweringlargereasoning}, where the accuracy for these tasks is evaluated by Qwen2.5-72B-Instruct~\citep{qwen2.5}. For AIME, we report average accuracy across five runs.

\paragraph{Models.}
We design our framework based on smolagents~\citep{smolagents}. For the main GAIA and WebWalkerQA experiments, we evaluate four model configurations: GPT-4o~\citep{openai2024gpt4o}, GPT-4.1~\citep{openai2025gpt4.1}, Claude 3.7 Sonnet~\citep{claude37sonnet}, and DeepSeek-V4-Pro~\citep{deepseekai2026deepseekv4}. For the IBIS benchmark, we additionally include Gemini-2.5-Pro~\citep{google2025gemini2.5-pro} and Qwen3-max~\citep{qwen3max}. For the more challenging BrowseComp and BrowseComp-zh evaluations, we report results with Claude Sonnet 4~\citep{anthropic2025claude-sonnet4}.

\paragraph{Baselines.}
We compare our method against a broad set of strong deep research baselines. The open-source baselines include smolagents DR~\citep{smolagents_open_deep_research}, BrowseMaster~\citep{pang2025browsemasterscalablewebbrowsing}, AWorld~\citep{xie2025profileawaremaneuveringdynamicmultiagent}, OWL~\citep{hu2025owloptimizedworkforcelearning}, OAgents~\citep{zhu2025oagentsempiricalstudybuilding}, Agent-KB~\citep{tang2025agentkbleveragingcrossdomain}, WebExplorer~\citep{liu2025webexplorerexploreevolvetraining}, MiroFlow~\citep{su2026miroflowhighperformancerobustopensource}, MemoBrain~\citep{qian2026memobrainexecutivememoryagentic}, Agentic Reasoning~\citep{wu2025agenticreasoningstreamlinedframework}. We also report closed-source systems, including OpenAI DR~\citep{openai2025introducing} and Metaso DR~\citep{metaso_deepresearch_2025}.

\subsection{Diagnosing Inertia Bias with IBIS}

\begin{figure}[t]
    \centering
    \includegraphics[width=\linewidth]{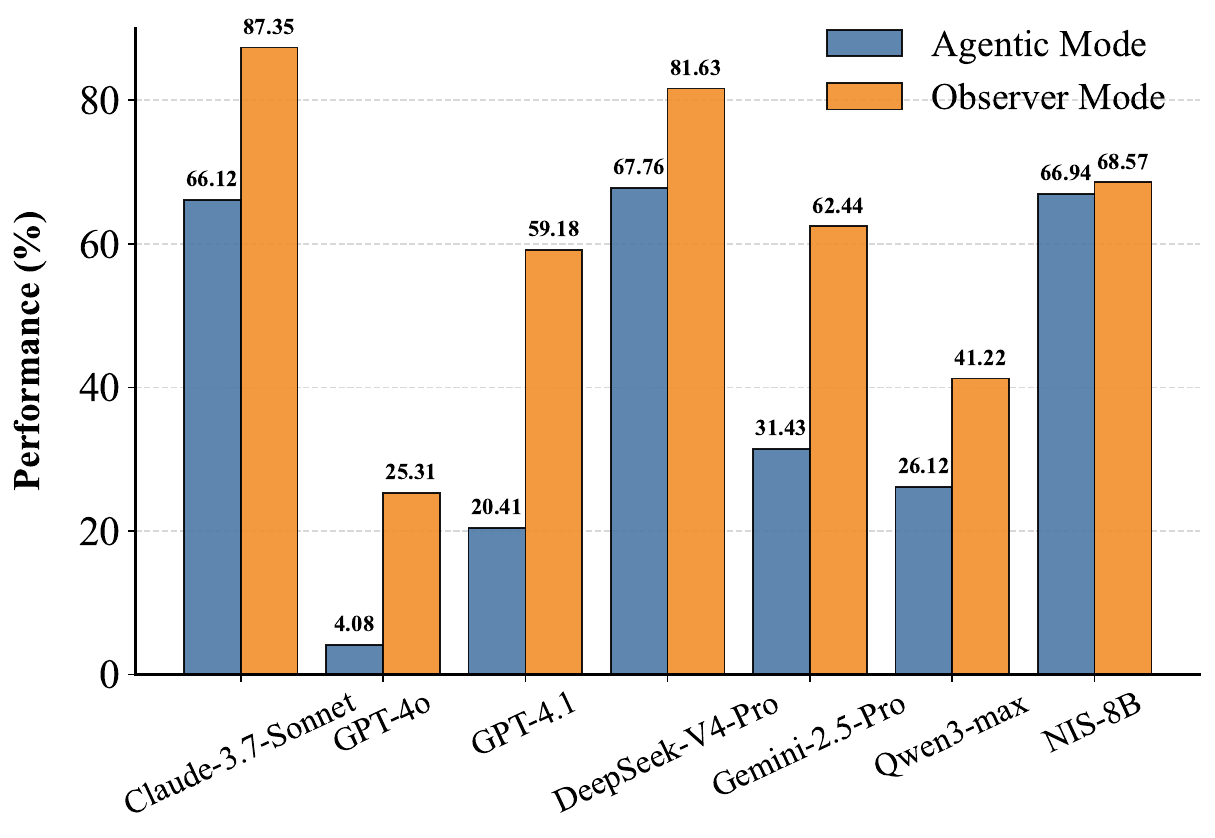} 
    \caption{Quantification of inertia bias on IBIS Should Re-search subset. Agentic Mode frames search results as self-initiated actions, while Observer Mode presents them as neutral external references.}
    \label{fig:ibis_results}
\end{figure}

We utilize our proposed IBIS benchmark to explicitly quantify inertia bias across different LLMs. As shown in Figure~\ref{fig:ibis_results}, switching from Agentic Mode to Observer Mode yields a performance improvement of approximately 15\% to 30\% across almost all models. This significant gap empirically confirms the prevalence of inertia bias in mainstream LLMs. Notably, even in Observer Mode, the success rate for rejecting irrelevant results rarely exceeds 90\%. This is primarily attributed to inherent model variance; as detailed in our analysis in Appendix~\ref{Apendix:IBIS_Experiments}, models like GPT-4o exhibit a strong inherent preference for calling \texttt{visit\_page} regardless of context. Furthermore, NIS-8B shows minimal performance gap between the two modes, demonstrating stronger immunity to inertia bias. Since Agentic and Observer Mode share identical search results, this gap could in principle still reflect an interaction with search-result content rather than action-history ownership; a paired flip analysis in Appendix~\ref{appendix:flip_analysis} rules this out, showing that shallow surface features of the search results cannot predict which samples flip between modes (AUC $\approx$ 0.50). For results on the \textit{Should Visit Page} subset and evaluation of our proposed Direct Mode, please refer to Appendix~\ref{Apendix:IBIS_Experiments}.

\subsection{Main Results in Deep Research}
\label{main_results}

\paragraph{Main Results.}
Table~\ref{tab:main_results} presents the main experimental results. Overall, NIS-Agent consistently outperforms existing workflow baselines. Specifically: (i) The comparisons using GPT-4.1 and GPT-4o provide direct evidence of our framework's efficacy. Under both model configurations, NIS-Agent achieves substantial improvements over the smolagents DR baseline on both benchmarks, with gains reaching approximately 8 to 11 percentage points. (ii) When powered by Claude-3.7-Sonnet, NIS-Agent achieves 72.73\% on GAIA and 68.50\% on WebWalkerQA, outperforming competing frameworks that rely on models of comparable or even greater capability, such as OWL (Claude-3.7-Sonnet), WebExplorer (Claude-Sonnet-4) and MiroFlow (GPT-5). (iii) Equipped with DeepSeek-V4-Pro, the most capable frontier model at the time of evaluation, NIS-Agent sets a new state-of-the-art among open-source frameworks on both benchmarks, achieving 81.21\% on GAIA and 75.00\% on WebWalkerQA, surpassing all compared baselines by a substantial margin. (iv) NIS-8B, trained as described in Section~\ref{sec:nis_8b}, remains competitive with the closed-source variants on both benchmarks: under the same NIS-Agent framework, its average GAIA performance is comparable to using GPT-4o as the backbone (63.64 vs. 61.82), showing that the model can be trained to be intrinsically less susceptible to inertia bias.

\paragraph{More Challenging Tasks.}
\begin{table}[t]
\centering
\small
\caption{Results on BrowseComp and BrowseComp-zh.}
\label{tab:browsecomp}
\resizebox{\linewidth}{!}{%
\begin{tabular}{l l c c}
\toprule
\textbf{Framework} & \textbf{Model} & \textbf{BC} & \textbf{BC-zh} \\
\midrule
OpenAI DR & o3-SFT & \textbf{51.5} & 42.9 \\
Metaso DR & -- & 12.0 & \underline{45.3} \\
Agentic Reasoning & Deepseek-R1 & 5.5 & 29.0 \\
WebExplorer & Claude-Sonnet-4 & 12.2 & 29.1 \\
\rowcolor[rgb]{0.902,1,0.902} NIS-Agent (Ours) & Claude-Sonnet-4 & \underline{25.0} & \textbf{45.9} \\
\bottomrule
\end{tabular}%
}
\end{table}

To further evaluate the robustness of NIS-Agent beyond GAIA and WebWalkerQA, we additionally test it on BrowseComp and BrowseComp-zh. Table~\ref{tab:browsecomp} shows that NIS-Agent achieves 25.0 on BrowseComp and 45.9 on BrowseComp-zh. These results substantially outperform the compared open-source baselines. Although OpenAI DR remains stronger on BrowseComp overall, NIS-Agent is highly competitive in this broader comparison and attains the best result among the listed systems on BrowseComp-zh.

\begin{table}[t]
  \centering
  \caption{Token usage on GAIA, measured as average tokens per query.}
  \label{tab:token-comparison}
  \resizebox{\linewidth}{!}{%
  \begin{tabular}{l c c c}
    \toprule
    \textbf{Method} & \textbf{Input} & \textbf{Output} & \textbf{Total} \\
    \midrule
    smolagents DR & 217.6k & 2.2k & 219.8k \\
    NIS-Agent & 144.1k \uagdown{33.8} & 3.2k \uagup{45.5} & 147.3k \uagdown{33.0} \\
    \bottomrule
  \end{tabular}}
\end{table}

\paragraph{Efficiency.}
As shown in Table~\ref{tab:token-comparison}, on GAIA with GPT-4o, the NIS-Agent reduces average total token usage per query by 33\% relative to smolagents DR, mainly by cutting input tokens.

\subsection{Ablation Study}
\begin{table}[t]
    \centering
    \caption{Ablation study of NIS-Agent components on GAIA with GPT-4o. Since the Stage 2 validation operates on the output of Stage 1, we stack the two stages along their dependency direction rather than removing them independently.}
    \label{tab:ablation}
    \resizebox{\linewidth}{!}{
    \begin{tabular}{lcccc}
        \toprule
        \multirow{2}{*}{\textbf{Method}} & \multirow{2}{*}{\textbf{Filter}} & \multicolumn{2}{c}{\textbf{Validation}} & \multirow{2}{*}{\textbf{Avg.}} \\
        \cmidrule(lr){3-4}
        & & \textbf{Stage 1} & \textbf{Stage 2} & \\
        \midrule
        NIS-Agent & \cmark & \cmark & \cmark & \textbf{61.82} \\
        \quad w/o Filter & \xmark & \cmark & \cmark & 59.39 \\
        \quad w/o Stage 2 & \cmark & \cmark & \xmark & 58.79 \\
        \quad w/o Validation & \cmark & \xmark & \xmark & 56.96 \\
        \bottomrule
    \end{tabular}
    }
\end{table}
\begin{table}[t]
\centering
\small
\caption{Transfer results of the isolation-based stepwise validation module on AIME. Results are Avg@5.}
\label{tab:aime_validation}
\resizebox{\linewidth}{!}{%
\begin{tabular}{l l c c}
\toprule
\textbf{Method} & \textbf{Model} & \textbf{AIME 2024} & \textbf{AIME 2025} \\
\midrule
CoT Validation & GPT-4o & 13.7 & 7.3 \\
\rowcolor[rgb]{0.902,1,0.902} Our Validation & GPT-4o & 29.4 & 26.7 \\
CoT Validation & GPT-4.1 & 49.3 & 35.3 \\
\rowcolor[rgb]{0.902,1,0.902} Our Validation & GPT-4.1 & \textbf{74.7} & \textbf{68.7} \\
\bottomrule
\end{tabular}%
}
\end{table}

We conduct ablation experiments to assess whether the gains of NIS-Agent indeed come from the two proposed context-isolation mechanisms. On the full GAIA validation set with GPT-4o as the backbone model, removing the context-isolated filter or the isolation-based stepwise validation module causes a clear performance drop, as shown in Table~\ref{tab:ablation}. Since the Stage 2 validation operates on the output of Stage 1, we cannot ablate it independently, so we instead stack the two stages along their dependency direction: adding Stage 1 alone already improves GAIA accuracy from 56.96 to 58.79, and further adding Stage 2 raises it to 61.82. Both modules and both validation stages thus contribute a distinct and non-redundant share of the overall gain. We report a qualitative analysis of the remaining failure cases, including cases where context isolation itself hurts performance, in Appendix~\ref{appendix:failure_analysis}.

We further study AIME 2024 and AIME 2025 by attaching our validation module to a ReAct pipeline. Both settings use the same problem-solving prompt and differ only in the validation procedure. As shown in Table~\ref{tab:aime_validation}, replacing a standard CoT-based validation scheme with our validation module yields substantial improvements for both GPT-4o and GPT-4.1. These results provide further evidence that inertia bias is an inherent weakness of LLMs in multi-step iterative reasoning. And our method generalizes beyond deep research tasks.

\section{Conclusions}

In this work, we identify \textbf{inertia bias}, the tendency of LLMs to reinforce their prior choices regardless of objective correctness, and quantify it with the IBIS benchmark. We show that this bias amplifies both search noise and contextual noise in deep research, and propose NIS-Agent to isolate the judgments most vulnerable to action-history ownership. We further train an 8B open-source model to be intrinsically more resistant to inertia bias, attaining performance comparable to GPT-4o.

\section{Limitations}
Our work has several limitations that suggest directions for future research.

\paragraph{Mechanistic Interpretability.}
While we have empirically identified and quantified inertia bias, our analysis currently treats the underlying Large Language Model primarily as a black box. We do not investigate the mechanistic origins of this phenomenon to determine whether it stems from specific attention patterns, pretraining data distributions, or artifacts of reinforcement learning alignment. A deeper understanding of the internal representations that drive this self-reinforcing tendency is necessary to effectively address the root cause of the bias.
\paragraph{Generalization across Domains.}
We validate inertia bias on two types of tasks: information seeking, where IBIS isolates the decision boundary between browsing a page and reformulating a query, and multi-step mathematical reasoning, where our validation module also improves performance on AIME 2024 and AIME 2025 (Table~\ref{tab:aime_validation}). We therefore do not claim that inertia bias is a fully general phenomenon across all agentic workflows; rather, we speculate that it may extend to other agentic scenarios, and we leave coding agents, embodied agents, and planning agents as future work.

\paragraph{Risks of Isolation and Validation Failures.}
NIS-Agent isolates and validates the model's own action history to counteract inertia bias, but this intervention is not guaranteed to be correct at every step. If the context-isolated filter mistakenly discards a relevant page, or if the isolation-based validation module endorses a flawed intermediate conclusion, the agent may still return an overconfident final answer, potentially citing sources that do not actually support it. In high-stakes information-seeking settings, such errors could mislead a user who trusts the agent's answer without independently verifying it. We therefore view NIS-Agent as a mitigation that reduces, rather than eliminates, the risk of self-reinforcing errors, and we recommend that deployments in high-stakes settings keep a human in the loop to verify cited evidence before acting on the agent's conclusions.

% \section{Ethics Statement}
% The datasets used in this study are publicly available and have been pre-anonymized. Furthermore, the IBIS benchmark proposed in this work does not contain any information that identifies individuals, nor does it include any offensive content. We have manually verified the data to ensure it adheres to ethical standards and privacy requirements.

% Regarding the preparation of the manuscript, Large Language Models (LLMs) were utilized solely for language polishing and grammatical improvements. The core content and original ideas were authored entirely by the researchers. We have carefully reviewed and verified all AI-assisted edits to ensure accuracy and to prevent any potential hallucinations or misinterpretations.

\section*{Acknowledgements}
This work was supported in part by The National Nature Science Foundation of China (Grant Nos.: 62303406, 62432014), in part by Ant Group, in part by Yongjiang Talent Introduction Programme (Grant No.: 2022A-240-G).

\bibliography{custom}

@misc{Significant_Gravitas_AutoGPT,
author = {{Significant Gravitas}},
year = {2023},
license = {MIT},
title = {{AutoGPT}},
url = {https://github.com/Significant-Gravitas/AutoGPT}
}

@inproceedings{yao2023react,
  title={React: Synergizing reasoning and acting in language models},
  author={Yao, Shunyu and Zhao, Jeffrey and Yu, Dian and Du, Nan and Shafran, Izhak and Narasimhan, Karthik and Cao, Yuan},
  booktitle={International Conference on Learning Representations (ICLR)},
  year={2023}
}

@inproceedings{NEURIPS2023_7a92bcde,
 author = {Valmeekam, Karthik and Marquez, Matthew and Olmo, Alberto and Sreedharan, Sarath and Kambhampati, Subbarao},
 booktitle = {Advances in Neural Information Processing Systems},
 editor = {A. Oh and T. Naumann and A. Globerson and K. Saenko and M. Hardt and S. Levine},
 pages = {38975--38987},
 publisher = {Curran Associates, Inc.},
 title = {PlanBench: An Extensible Benchmark for Evaluating Large Language Models on Planning and Reasoning about Change},
 url = {https://proceedings.neurips.cc/paper_files/paper/2023/file/7a92bcdede88c7afd108072faf5485c8-Paper-Datasets_and_Benchmarks.pdf},
 volume = {36},
 year = {2023}
}

@misc{packer2024memgptllmsoperatingsystems,
      title={MemGPT: Towards LLMs as Operating Systems}, 
      author={Charles Packer and Sarah Wooders and Kevin Lin and Vivian Fang and Shishir G. Patil and Ion Stoica and Joseph E. Gonzalez},
      year={2024},
      eprint={2310.08560},
      archivePrefix={arXiv},
      primaryClass={cs.AI},
      url={https://arxiv.org/abs/2310.08560}, 
}

@misc{xu2025amemagenticmemoryllm,
      title={A-MEM: Agentic Memory for LLM Agents}, 
      author={Wujiang Xu and Kai Mei and Hang Gao and Juntao Tan and Zujie Liang and Yongfeng Zhang},
      year={2025},
      eprint={2502.12110},
      archivePrefix={arXiv},
      primaryClass={cs.CL},
      url={https://arxiv.org/abs/2502.12110}, 
}

@inproceedings{zhu-etal-2025-knowagent,
    title = "{K}now{A}gent: Knowledge-Augmented Planning for {LLM}-Based Agents",
    author = "Zhu, Yuqi  and
      Qiao, Shuofei  and
      Ou, Yixin  and
      Deng, Shumin  and
      Lyu, Shiwei  and
      Shen, Yue  and
      Liang, Lei  and
      Gu, Jinjie  and
      Chen, Huajun  and
      Zhang, Ningyu",
    editor = "Chiruzzo, Luis  and
      Ritter, Alan  and
      Wang, Lu",
    booktitle = "Findings of the Association for Computational Linguistics: NAACL 2025",
    month = apr,
    year = "2025",
    address = "Albuquerque, New Mexico",
    publisher = "Association for Computational Linguistics",
    url = "https://aclanthology.org/2025.findings-naacl.205/",
    doi = "10.18653/v1/2025.findings-naacl.205",
    pages = "3709--3732",
    ISBN = "979-8-89176-195-7"
}

@misc{wang2024agentworkflowmemory,
      title={Agent Workflow Memory}, 
      author={Zora Zhiruo Wang and Jiayuan Mao and Daniel Fried and Graham Neubig},
      year={2024},
      eprint={2409.07429},
      archivePrefix={arXiv},
      primaryClass={cs.CL},
      url={https://arxiv.org/abs/2409.07429}, 
}

@misc{tang2025agentkbleveragingcrossdomain,
      title={Agent KB: Leveraging Cross-Domain Experience for Agentic Problem Solving}, 
      author={Xiangru Tang and Tianrui Qin and Tianhao Peng and Ziyang Zhou and Daniel Shao and Tingting Du and Xinming Wei and Peng Xia and Fang Wu and He Zhu and Ge Zhang and Jiaheng Liu and Xingyao Wang and Sirui Hong and Chenglin Wu and Hao Cheng and Chi Wang and Wangchunshu Zhou},
      year={2025},
      eprint={2507.06229},
      archivePrefix={arXiv},
      primaryClass={cs.CL},
      url={https://arxiv.org/abs/2507.06229}, 
}

@misc{java2025characterizingdeepresearchbenchmark,
      title={Characterizing Deep Research: A Benchmark and Formal Definition}, 
      author={Abhinav Java and Ashmit Khandelwal and Sukruta Midigeshi and Aaron Halfaker and Amit Deshpande and Navin Goyal and Ankur Gupta and Nagarajan Natarajan and Amit Sharma},
      year={2025},
      eprint={2508.04183},
      archivePrefix={arXiv},
      primaryClass={cs.CL},
      url={https://arxiv.org/abs/2508.04183}, 
}

@Misc{smolagents,
  title =        {`smolagents`: a smol library to build great agentic systems.},
  author =       {Aymeric Roucher and Albert Villanova del Moral and Thomas Wolf and Leandro von Werra and Erik Kaunismäki},
  howpublished = {\url{https://github.com/huggingface/smolagents}},
  year =         {2025}
}

@misc{smolagents_open_deep_research,
  author       = {Aymeric Roucher and Albert Villanova del Moral and Thomas Wolf and Leandro von Werra and Erik Kaunismäki},
  title        = {Open-source DeepResearch – Freeing our search agents},
  howpublished = {\url{https://huggingface.co/blog/open-deep-research}},
  month        = feb,
  year         = {2025},
}

@misc{li2025webthinkerempoweringlargereasoning,
      title={WebThinker: Empowering Large Reasoning Models with Deep Research Capability}, 
      author={Xiaoxi Li and Jiajie Jin and Guanting Dong and Hongjin Qian and Yutao Zhu and Yongkang Wu and Ji-Rong Wen and Zhicheng Dou},
      year={2025},
      eprint={2504.21776},
      archivePrefix={arXiv},
      primaryClass={cs.CL},
      url={https://arxiv.org/abs/2504.21776}, 
}

@misc{tao2025webshaperagenticallydatasynthesizing,
      title={WebShaper: Agentically Data Synthesizing via Information-Seeking Formalization}, 
      author={Zhengwei Tao and Jialong Wu and Wenbiao Yin and Junkai Zhang and Baixuan Li and Haiyang Shen and Kuan Li and Liwen Zhang and Xinyu Wang and Yong Jiang and Pengjun Xie and Fei Huang and Jingren Zhou},
      year={2025},
      eprint={2507.15061},
      archivePrefix={arXiv},
      primaryClass={cs.CL},
      url={https://arxiv.org/abs/2507.15061}, 
}

@misc{mialon2023gaiabenchmarkgeneralai,
      title={GAIA: a benchmark for General AI Assistants}, 
      author={Grégoire Mialon and Clémentine Fourrier and Craig Swift and Thomas Wolf and Yann LeCun and Thomas Scialom},
      year={2023},
      eprint={2311.12983},
      archivePrefix={arXiv},
      primaryClass={cs.CL},
      url={https://arxiv.org/abs/2311.12983}, 
}

@misc{wei2025browsecompsimplechallengingbenchmark,
      title={BrowseComp: A Simple Yet Challenging Benchmark for Browsing Agents}, 
      author={Jason Wei and Zhiqing Sun and Spencer Papay and Scott McKinney and Jeffrey Han and Isa Fulford and Hyung Won Chung and Alex Tachard Passos and William Fedus and Amelia Glaese},
      year={2025},
      eprint={2504.12516},
      archivePrefix={arXiv},
      primaryClass={cs.CL},
      url={https://arxiv.org/abs/2504.12516}, 
}

@misc{zhou2025browsecompzhbenchmarkingwebbrowsing,
      title={BrowseComp-ZH: Benchmarking Web Browsing Ability of Large Language Models in Chinese}, 
      author={Peilin Zhou and Bruce Leon and Xiang Ying and Can Zhang and Yifan Shao and Qichen Ye and Dading Chong and Zhiling Jin and Chenxuan Xie and Meng Cao and Yuxin Gu and Sixin Hong and Jing Ren and Jian Chen and Chao Liu and Yining Hua},
      year={2025},
      eprint={2504.19314},
      archivePrefix={arXiv},
      primaryClass={cs.CL},
      url={https://arxiv.org/abs/2504.19314}, 
}

@misc{wu2025webwalkerbenchmarkingllmsweb,
      title={WebWalker: Benchmarking LLMs in Web Traversal}, 
      author={Jialong Wu and Wenbiao Yin and Yong Jiang and Zhenglin Wang and Zekun Xi and Runnan Fang and Linhai Zhang and Yulan He and Deyu Zhou and Pengjun Xie and Fei Huang},
      year={2025},
      eprint={2501.07572},
      archivePrefix={arXiv},
      primaryClass={cs.CL},
      url={https://arxiv.org/abs/2501.07572}, 
}

@misc{zhu2025oagentsempiricalstudybuilding,
      title={OAgents: An Empirical Study of Building Effective Agents}, 
      author={He Zhu and Tianrui Qin and King Zhu and Heyuan Huang and Yeyi Guan and Jinxiang Xia and Yi Yao and Hanhao Li and Ningning Wang and Pai Liu and Tianhao Peng and Xin Gui and Xiaowan Li and Yuhui Liu and Yuchen Eleanor Jiang and Jun Wang and Changwang Zhang and Xiangru Tang and Ge Zhang and Jian Yang and Minghao Liu and Xitong Gao and Jiaheng Liu and Wangchunshu Zhou},
      year={2025},
      eprint={2506.15741},
      archivePrefix={arXiv},
      primaryClass={cs.AI},
      url={https://arxiv.org/abs/2506.15741}, 
}

@misc{pang2025browsemasterscalablewebbrowsing,
      title={BrowseMaster: Towards Scalable Web Browsing via Tool-Augmented Programmatic Agent Pair}, 
      author={Xianghe Pang and Shuo Tang and Rui Ye and Yuwen Du and Yaxin Du and Siheng Chen},
      year={2025},
      eprint={2508.09129},
      archivePrefix={arXiv},
      primaryClass={cs.AI},
      url={https://arxiv.org/abs/2508.09129}, 
}

@misc{guo2024largelanguagemodelbased,
      title={Large Language Model based Multi-Agents: A Survey of Progress and Challenges}, 
      author={Taicheng Guo and Xiuying Chen and Yaqi Wang and Ruidi Chang and Shichao Pei and Nitesh V. Chawla and Olaf Wiest and Xiangliang Zhang},
      year={2024},
      eprint={2402.01680},
      archivePrefix={arXiv},
      primaryClass={cs.CL},
      url={https://arxiv.org/abs/2402.01680}, 
}

@misc{openai2025introducing,
  title        = {Introducing Deep Research},
  author       = {{OpenAI}},
  year         = {2025},
  month        = {February 2},
  url          = {https://openai.com/index/introducing-deep-research/},
}

@misc{li2025websailornavigatingsuperhumanreasoning,
      title={WebSailor: Navigating Super-human Reasoning for Web Agent}, 
      author={Kuan Li and Zhongwang Zhang and Huifeng Yin and Liwen Zhang and Litu Ou and Jialong Wu and Wenbiao Yin and Baixuan Li and Zhengwei Tao and Xinyu Wang and Weizhou Shen and Junkai Zhang and Dingchu Zhang and Xixi Wu and Yong Jiang and Ming Yan and Pengjun Xie and Fei Huang and Jingren Zhou},
      year={2025},
      eprint={2507.02592},
      archivePrefix={arXiv},
      primaryClass={cs.CL},
      url={https://arxiv.org/abs/2507.02592}, 
}

@misc{hu2025owloptimizedworkforcelearning,
      title={OWL: Optimized Workforce Learning for General Multi-Agent Assistance in Real-World Task Automation}, 
      author={Mengkang Hu and Yuhang Zhou and Wendong Fan and Yuzhou Nie and Bowei Xia and Tao Sun and Ziyu Ye and Zhaoxuan Jin and Yingru Li and Qiguang Chen and Zeyu Zhang and Yifeng Wang and Qianshuo Ye and Bernard Ghanem and Ping Luo and Guohao Li},
      year={2025},
      eprint={2505.23885},
      archivePrefix={arXiv},
      primaryClass={cs.AI},
      url={https://arxiv.org/abs/2505.23885}, 
}

@misc{liu2025webexplorerexploreevolvetraining,
      title={WebExplorer: Explore and Evolve for Training Long-Horizon Web Agents}, 
      author={Junteng Liu and Yunji Li and Chi Zhang and Jingyang Li and Aili Chen and Ke Ji and Weiyu Cheng and Zijia Wu and Chengyu Du and Qidi Xu and Jiayuan Song and Zhengmao Zhu and Wenhu Chen and Pengyu Zhao and Junxian He},
      year={2025},
      eprint={2509.06501},
      archivePrefix={arXiv},
      primaryClass={cs.CL},
      url={https://arxiv.org/abs/2509.06501}, 
}

@misc{openai2025gpt4.1,
  title        = {Introducing GPT-4.1 in the API},
  author       = {OpenAI},
  year         = {2025},
  howpublished = {\url{https://openai.com/index/gpt-4-1/}},
}

@misc{openai2024gpt4o,
  title        = {Hello GPT-4o — OpenAI},
  author       = {OpenAI},
  year         = {2024},
  howpublished = {\url{https://openai.com/index/hello-gpt-4o/}},
}

@misc{anthropic2025claude-sonnet4,
  title        = {Claude Sonnet 4},
  author       = {Anthropic},
  year         = {2025},
  howpublished = {\url{https://www.anthropic.com/claude/sonnet}},
}

@misc{google2025gemini2.5-pro,
  title        = {Gemini 2.5 Pro},
  author       = {Google / DeepMind},
  year         = {2025},
  howpublished = {\url{https://cloud.google.com/vertex-ai/generative-ai/docs/models/gemini/2-5-pro}},
}

@misc{xie2025profileawaremaneuveringdynamicmultiagent,
      title={Profile-Aware Maneuvering: A Dynamic Multi-Agent System for Robust GAIA Problem Solving by AWorld}, 
      author={Zhitian Xie and Qintong Wu and Chengyue Yu and Chenyi Zhuang and Jinjie Gu},
      year={2025},
      eprint={2508.09889},
      archivePrefix={arXiv},
      primaryClass={cs.AI},
      url={https://arxiv.org/abs/2508.09889}, 
}

@misc{zhang2025ascotadaptiveselfcorrectionchainofthought,
      title={ASCoT: An Adaptive Self-Correction Chain-of-Thought Method for Late-Stage Fragility in LLMs}, 
      author={Dongxu Zhang and Ning Yang and Jihua Zhu and Jinnan Yang and Miao Xin and Baoliang Tian},
      year={2025},
      eprint={2508.05282},
      archivePrefix={arXiv},
      primaryClass={cs.CL},
      url={https://arxiv.org/abs/2508.05282}, 
}

@misc{gan2025rethinkingexternalslowthinkingsnowball,
      title={Rethinking External Slow-Thinking: From Snowball Errors to Probability of Correct Reasoning}, 
      author={Zeyu Gan and Yun Liao and Yong Liu},
      year={2025},
      eprint={2501.15602},
      archivePrefix={arXiv},
      primaryClass={cs.AI},
      url={https://arxiv.org/abs/2501.15602}, 
}

@misc{zhang2024chainagentslargelanguage,
      title={Chain of Agents: Large Language Models Collaborating on Long-Context Tasks}, 
      author={Yusen Zhang and Ruoxi Sun and Yanfei Chen and Tomas Pfister and Rui Zhang and Sercan Ö. Arik},
      year={2024},
      eprint={2406.02818},
      archivePrefix={arXiv},
      primaryClass={cs.CL},
      url={https://arxiv.org/abs/2406.02818}, 
}

@misc{claude37sonnet,
  title        = {Claude 3.7 Sonnet and Claude Code},
  author       = {Anthropic},
  month        = {February},
  day          = {24},
  year         = {2025},
  url          = {https://www.anthropic.com/news/claude-3-7-sonnet},
}

@misc{qwen2.5,
    title = {Qwen2.5: A Party of Foundation Models},
    url = {https://qwenlm.github.io/blog/qwen2.5/},
    author = {Qwen Team},
    month = {September},
    year = {2024}
}

@misc{qwen3max,
    title = {Qwen3-Max: Just Scale it},
    author = {Qwen Team},
    month = {September},
    year = {2025}
}

@misc{qwen3technicalreport,
      title={Qwen3 Technical Report}, 
      author={Qwen Team},
      year={2025},
      eprint={2505.09388},
      archivePrefix={arXiv},
      primaryClass={cs.CL},
      url={https://arxiv.org/abs/2505.09388}, 
}

@misc{perez2022discoveringlanguagemodelbehaviors,
      title={Discovering Language Model Behaviors with Model-Written Evaluations}, 
      author={Ethan Perez and Sam Ringer and Kamilė Lukošiūtė and Karina Nguyen and Edwin Chen and Scott Heiner and Craig Pettit and Catherine Olsson and Sandipan Kundu and Saurav Kadavath and Andy Jones and Anna Chen and Ben Mann and Brian Israel and Bryan Seethor and Cameron McKinnon and Christopher Olah and Da Yan and Daniela Amodei and Dario Amodei and Dawn Drain and Dustin Li and Eli Tran-Johnson and Guro Khundadze and Jackson Kernion and James Landis and Jamie Kerr and Jared Mueller and Jeeyoon Hyun and Joshua Landau and Kamal Ndousse and Landon Goldberg and Liane Lovitt and Martin Lucas and Michael Sellitto and Miranda Zhang and Neerav Kingsland and Nelson Elhage and Nicholas Joseph and Noemí Mercado and Nova DasSarma and Oliver Rausch and Robin Larson and Sam McCandlish and Scott Johnston and Shauna Kravec and Sheer El Showk and Tamera Lanham and Timothy Telleen-Lawton and Tom Brown and Tom Henighan and Tristan Hume and Yuntao Bai and Zac Hatfield-Dodds and Jack Clark and Samuel R. Bowman and Amanda Askell and Roger Grosse and Danny Hernandez and Deep Ganguli and Evan Hubinger and Nicholas Schiefer and Jared Kaplan},
      year={2022},
      eprint={2212.09251},
      archivePrefix={arXiv},
      primaryClass={cs.CL},
      url={https://arxiv.org/abs/2212.09251}, 
}

@misc{sharma2025understandingsycophancylanguagemodels,
      title={Towards Understanding Sycophancy in Language Models}, 
      author={Mrinank Sharma and Meg Tong and Tomasz Korbak and David Duvenaud and Amanda Askell and Samuel R. Bowman and Newton Cheng and Esin Durmus and Zac Hatfield-Dodds and Scott R. Johnston and Shauna Kravec and Timothy Maxwell and Sam McCandlish and Kamal Ndousse and Oliver Rausch and Nicholas Schiefer and Da Yan and Miranda Zhang and Ethan Perez},
      year={2025},
      eprint={2310.13548},
      archivePrefix={arXiv},
      primaryClass={cs.CL},
      url={https://arxiv.org/abs/2310.13548}, 
}

@misc{wan2025unveilingconfirmationbiaschainofthought,
      title={Unveiling Confirmation Bias in Chain-of-Thought Reasoning}, 
      author={Yue Wan and Xiaowei Jia and Xiang Lorraine Li},
      year={2025},
      eprint={2506.12301},
      archivePrefix={arXiv},
      primaryClass={cs.LG},
      url={https://arxiv.org/abs/2506.12301}, 
}

@misc{huang2024largelanguagemodelsselfcorrect,
      title={Large Language Models Cannot Self-Correct Reasoning Yet}, 
      author={Jie Huang and Xinyun Chen and Swaroop Mishra and Huaixiu Steven Zheng and Adams Wei Yu and Xinying Song and Denny Zhou},
      year={2024},
      eprint={2310.01798},
      archivePrefix={arXiv},
      primaryClass={cs.CL},
      url={https://arxiv.org/abs/2310.01798}, 
}

@misc{shinn2023reflexionlanguageagentsverbal,
      title={Reflexion: Language Agents with Verbal Reinforcement Learning}, 
      author={Noah Shinn and Federico Cassano and Edward Berman and Ashwin Gopinath and Karthik Narasimhan and Shunyu Yao},
      year={2023},
      eprint={2303.11366},
      archivePrefix={arXiv},
      primaryClass={cs.AI},
      url={https://arxiv.org/abs/2303.11366}, 
}

@misc{gou2024criticlargelanguagemodels,
      title={CRITIC: Large Language Models Can Self-Correct with Tool-Interactive Critiquing}, 
      author={Zhibin Gou and Zhihong Shao and Yeyun Gong and Yelong Shen and Yujiu Yang and Nan Duan and Weizhu Chen},
      year={2024},
      eprint={2305.11738},
      archivePrefix={arXiv},
      primaryClass={cs.CL},
      url={https://arxiv.org/abs/2305.11738}, 
}

@misc{aime2024,
  title = {American Invitational Mathematics Examination 2024},
  author = {{MAA}},
  year = {2024},
  howpublished = {\url{https://maa.org}},
  note = {AIME contest problems}
}

@misc{aime2025,
  title = {American Invitational Mathematics Examination 2025},
  author = {{MAA}},
  year = {2025},
  howpublished = {\url{https://maa.org}},
  note = {AIME contest problems}
}

@misc{wu2025agenticreasoningstreamlinedframework,
      title={Agentic Reasoning: A Streamlined Framework for Enhancing LLM Reasoning with Agentic Tools}, 
      author={Junde Wu and Jiayuan Zhu and Yuyuan Liu and Min Xu and Yueming Jin},
      year={2025},
      eprint={2502.04644},
      archivePrefix={arXiv},
      primaryClass={cs.AI},
      url={https://arxiv.org/abs/2502.04644}, 
}

@misc{metaso_deepresearch_2025,
  title        = {Metaso DeepResearch},
  author       = {{Metaso}},
  year         = {2025},
  howpublished = {\url{https://metaso.cn/}},
  note         = {Accessed: 2026-04}
}

@misc{liu2024apigenautomatedpipelinegenerating,
      title={APIGen: Automated Pipeline for Generating Verifiable and Diverse Function-Calling Datasets}, 
      author={Zuxin Liu and Thai Hoang and Jianguo Zhang and Ming Zhu and Tian Lan and Shirley Kokane and Juntao Tan and Weiran Yao and Zhiwei Liu and Yihao Feng and Rithesh Murthy and Liangwei Yang and Silvio Savarese and Juan Carlos Niebles and Huan Wang and Shelby Heinecke and Caiming Xiong},
      year={2024},
      eprint={2406.18518},
      archivePrefix={arXiv},
      primaryClass={cs.CL},
      url={https://arxiv.org/abs/2406.18518}, 
}

@misc{shao2024deepseekmathpushinglimitsmathematical,
      title={DeepSeekMath: Pushing the Limits of Mathematical Reasoning in Open Language Models}, 
      author={Zhihong Shao and Peiyi Wang and Qihao Zhu and Runxin Xu and Junxiao Song and Xiao Bi and Haowei Zhang and Mingchuan Zhang and Y. K. Li and Y. Wu and Daya Guo},
      year={2024},
      eprint={2402.03300},
      archivePrefix={arXiv},
      primaryClass={cs.CL},
      url={https://arxiv.org/abs/2402.03300}, 
}

@misc{su2026miroflowhighperformancerobustopensource,
      title={MiroFlow: Towards High-Performance and Robust Open-Source Agent Framework for General Deep Research Tasks}, 
      author={Shiqian Su and Sen Xing and Xuan Dong and Muyan Zhong and Bin Wang and Xizhou Zhu and Yuntao Chen and Wenhai Wang and Yue Deng and Pengxiang Zhu and Ziyuan Liu and Tiantong Li and Jiaheng Yu and Zhe Chen and Lidong Bing and Jifeng Dai},
      year={2026},
      eprint={2602.22808},
      archivePrefix={arXiv},
      primaryClass={cs.AI},
      url={https://arxiv.org/abs/2602.22808}, 
}

@misc{deepseekai2026deepseekv4,
      title={DeepSeek-V4: Towards Highly Efficient Million-Token Context Intelligence},
      author={DeepSeek-AI},
      year={2026},
}

@misc{chen2026searchmorethinkless,
      title={Search More, Think Less: Rethinking Long-Horizon Agentic Search for Efficiency and Generalization}, 
      author={Qianben Chen and Tianrui Qin and King Zhu and Qiexiang Wang and Chengjun Yu and Shu Xu and Jiaqi Wu and Jiayu Zhang and Xinpeng Liu and Xin Gui and Jingyi Cao and Piaohong Wang and Dingfeng Shi and He Zhu and Tiannan Wang and Yuqing Wang and Maojia Song and Tianyu Zheng and Ge Zhang and Jian Yang and Jiaheng Liu and Minghao Liu and Yuchen Eleanor Jiang and Wangchunshu Zhou},
      year={2026},
      eprint={2602.22675},
      archivePrefix={arXiv},
      primaryClass={cs.CL},
      url={https://arxiv.org/abs/2602.22675}, 
}

@misc{qian2026memobrainexecutivememoryagentic,
      title={MemoBrain: Executive Memory as an Agentic Brain for Reasoning}, 
      author={Hongjin Qian and Zhao Cao and Zheng Liu},
      year={2026},
      eprint={2601.08079},
      archivePrefix={arXiv},
      primaryClass={cs.AI},
      url={https://arxiv.org/abs/2601.08079}, 
}

\appendix

\section{Additional Analysis of Inertia Bias in Web Search Agents}
\label{inertia bias}

\subsection{The Phenomenon of Inertia Bias in the Search Process}

In our agent search pipeline, we observe a discrepancy between the ideal 
and the actual behavior of the LLM. Ideally, the model should stop the 
search process once it determines that all retrieved results are irrelevant. 
However, when the agent design incorporates the full history of previous 
steps (including the LLM’s own queries), the model tends to continue 
selecting URLs even though it has already recognized them as irrelevant. 
This behavior reflects a form of path dependence or over-commitment.
\begin{itemize}
  \item \textbf{Step 1: Query Generation} -- The LLM produces a search query.
  \item \textbf{Step 2: Result Retrieval} -- A search tool returns URLs and summaries.
  \item \textbf{Step 3 (Ideal)}: If all results are irrelevant, stop the search.
  \item \textbf{Observed Phenomenon}: When the agent includes its full history, 
        the LLM often continues selecting URLs even though they appear irrelevant.
\end{itemize}

\subsection{Quantitative Analysis of Inertia Bias.}
Due to the iterative self-correcting capabilities inherent in autonomous agents, a failure during an initial search or browsing session does not necessarily preclude the possibility of recovering the correct path in subsequent steps. Consequently, relying solely on final task outcomes makes it difficult to quantitatively isolate task failures directly attributable to search noise caused by inertia bias. However, the negative impact of this bias is objectively real and measurable through process inefficiencies, specifically in the form of wasted reasoning steps and erroneous intermediate answers. Since current LLM-based automated evaluation methods are neither sufficiently realistic nor reliable enough to distinguish these subtle noise patterns from legitimate reasoning processes, we opted for a rigorous manual evaluation strategy. 

To provide a precise quantitative assessment of how inertia bias affects performance, we conducted a manual statistical analysis focusing on errors induced by search noise and contextual noise. This study was executed using the smolagents + GPT-4.1 framework on the GAIA benchmark.

\begin{figure}[h]
\centering
    \includegraphics[width=\columnwidth]{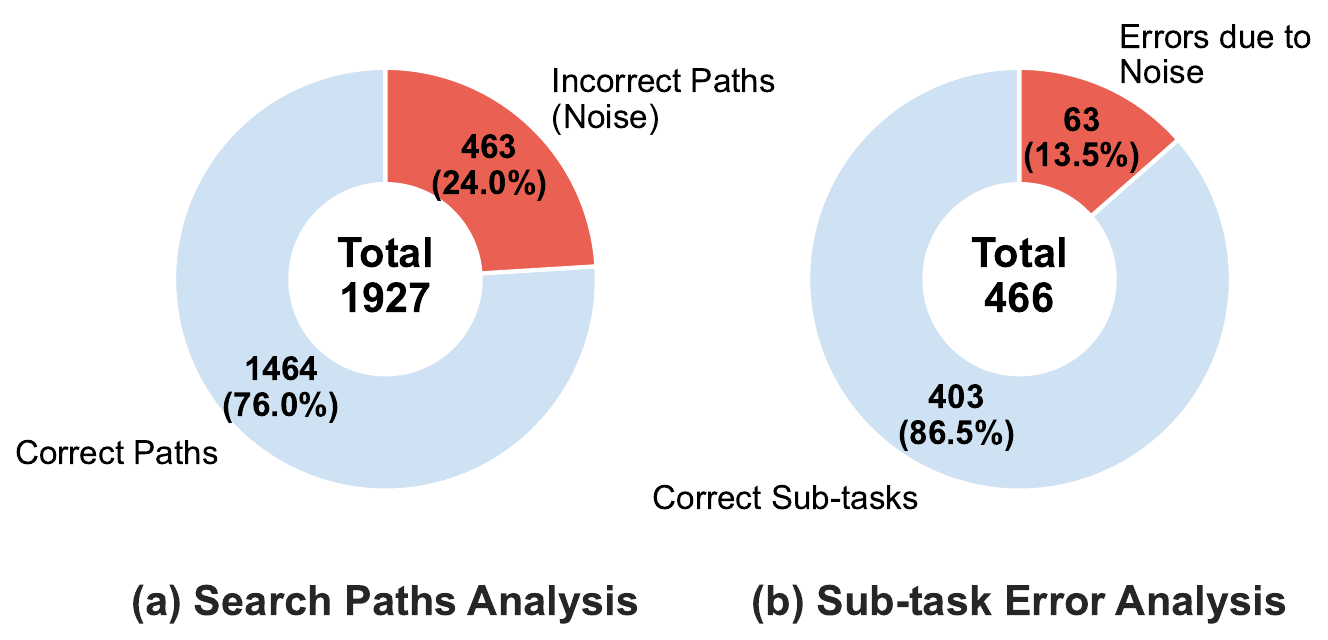}
    \caption{Analysis of search noise on GAIA. (a) Distribution of search paths, showing the proportion of incorrect paths caused by search noise. (b) Impact on sub-tasks, illustrating how search noise leads to erroneous conclusions.}
    \label{fig:quantity_search_noise}
\end{figure}

\paragraph{Impact of Search Noise.} We analyzed the search noise resulting from inertia bias by recording the following metrics:
(1) Total Searches: The total number of search actions performed.
(2) Incorrect Paths due to Search Noise: The number of times search noise was not correctly identified, resulting in the agent following an incorrect path.
(3) Total Sub-tasks: The total number of sub-tasks assigned to the SearchAgent (focusing on sub-tasks rather than macro-tasks, as each search is specific to a sub-task).
(4) Sub-task Errors due to Search Noise: The number of times failing to identify search noise and following an incorrect path directly led to an erroneous conclusion for that sub-task.
The results in Figure~\ref{fig:quantity_search_noise} indicate that search noise substantially impacts search efficiency, affecting 24.0\% of total searches. Furthermore, although the agent can potentially compensate for invalid searches in subsequent iterations, the final conclusions of 13.5\%  of the sub-tasks are still adversely affected by this noise.

\begin{figure}[h]
\centering
    \includegraphics[width=.5\columnwidth]{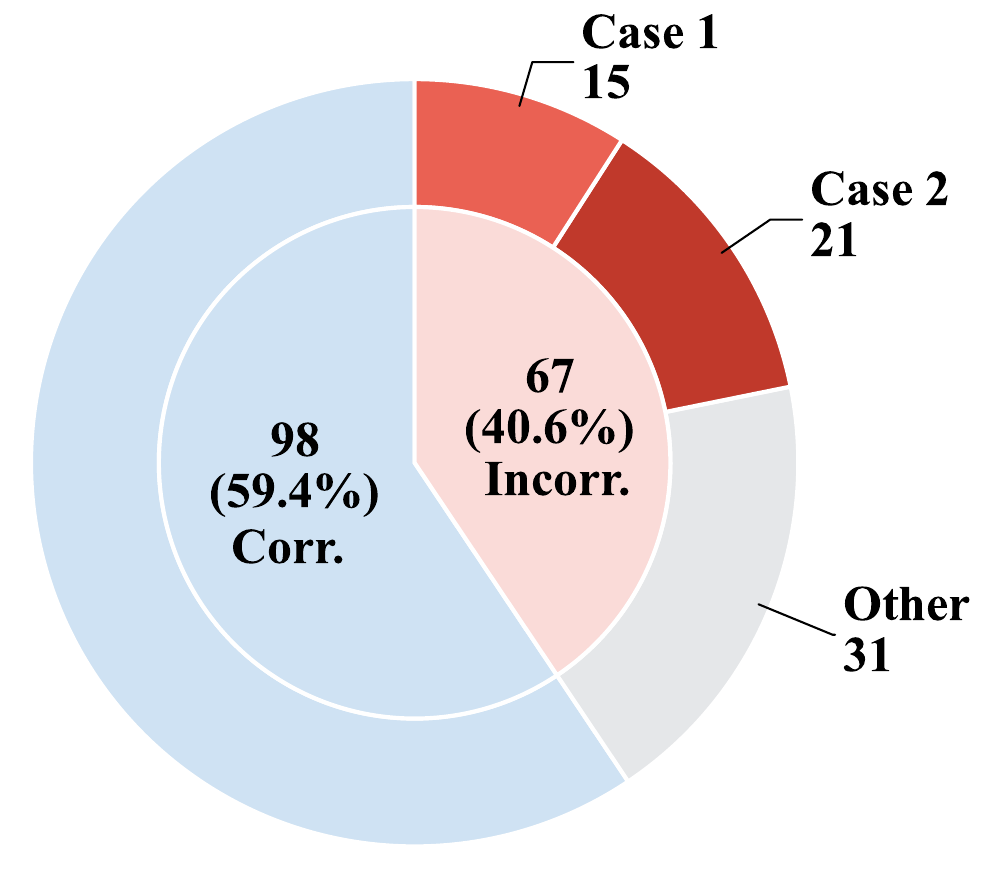}
    \caption{Breakdown of errors by contextual noise on GAIA. The inner ring shows correct vs. incorrect answers, while the outer ring details the composition of errors: Case 1 (misinterpretation with complete observations) and Case 2 (premature termination with partial observations).}
    \label{fig:quantity_contextual_noise}
\end{figure}

\paragraph{Impact of Contextual Noise.} We further analyzed errors caused by context bias (noise) by categorizing them into two distinct cases:
Case 1: The context contained all observations necessary to derive the final answer, and these observations were correct; however, the ManagerAgent reached an incorrect conclusion due to interference from other irrelevant information.
Case 2: The context contained only partial observations necessary for the final answer (which were correct); however, due to interference from other information, the ManagerAgent prematurely concluded that sufficient information had been obtained, leading to an erroneous conclusion.
As shown in Figure~\ref{fig:quantity_contextual_noise}, contextual noise significantly influences the final conclusion. Among the incorrect answers, a notable portion resulted specifically from the agent's inability to filter out contextual noise, either leading to misinterpretation of full information (Case 1) or premature termination based on partial information (Case 2).

\subsection{Conceptual Differentiation: Inertia Bias versus Related Biases}
\label{appendix:inertia_vs_sycophancy}

We place the conceptual differentiation in the appendix because the main text focuses on the empirical diagnosis and mitigation of inertia bias, while this section clarifies its boundary relative to adjacent bias categories.

\paragraph{Distinct triggering mechanism (vs. sycophancy).}
Sycophancy typically describes an LLM's tendency to align with external user inputs or perceived human preferences, even when such alignment conflicts with facts~\citep{perez2022discoveringlanguagemodelbehaviors, sharma2025understandingsycophancylanguagemodels}. Inertia bias is entirely internal. The model is not attempting to please the user; instead, it becomes irrationally anchored to its own previously generated action, such as a query, plan, or intermediate conclusion. One may loosely describe this as a form of ``self-sycophancy,'' but the operational trigger is different: the pressure comes from self-authored action history rather than from external human framing.

\paragraph{Action-space focus (vs. confirmation bias).}
Confirmation bias usually concerns commitment to a prior belief or hypothesis, leading the model to favor evidence that supports that belief. Inertia bias, by contrast, is centered on the execution trajectory of the agent. It manifests as persistence in a flawed tool-use path or plan, even when new observations already indicate that the current trajectory should be revised. In this sense, the object of commitment is not merely a proposition, but the agent's own prior action.

\paragraph{Behavioral measurability through IBIS.}
A further distinction of inertia bias in our work is that it is made behaviorally measurable. IBIS keeps the task and search observations fixed while manipulating whether the model ``owns'' the prior search action. This controlled design isolates the effect of action-history ownership on the next-step decision, enabling a more direct diagnosis of inertia bias than broad error-rate comparisons alone.

\subsection{Annotation Protocol and Quality Control}
\label{appendix:annotation_protocol}

IBIS is not constructed in a single pass. We first prompt qwen3-max to generate 1,016 candidate samples spanning domains such as health, economy, technology, climate, and basic science. Three independent large models, namely Gemini-2.5-Pro, DeepSeek-R1, and Claude-3.7-Sonnet, then score each candidate along three dimensions: the factual verifiability of the question, the uniqueness of the answer, and the consistency between the query and the information need. Candidates that do not meet a minimum quality threshold on all three dimensions are discarded. This filtering step leaves 712 high-quality samples.

For each retained sample, we execute its query against Google Search. Three human annotators, acting as a neutral third party, then independently categorize the result based on the retrieved web abstract. Each sample is assigned to one of three classes. The first class is Should Re-search, where the abstracts are clearly irrelevant to the information need. The second class is Should Visit Page, where at least one result appears relevant based on its title and snippet. The third class is Should Return Answer, where the abstract already contains the answer, so visiting a page is unnecessary. A sample is retained only if all three annotators assign the same category. Any disagreement leads to exclusion rather than adjudication, so annotation quality and consistency are controlled without an additional arbitration step that could itself introduce bias. The resulting inter-annotator agreement, measured by Fleiss' $\kappa$ across the three categories, is 0.684, indicating substantial agreement. IBIS focuses on the decision point most directly tied to inertia bias, namely whether to continue or abandon the current search path. We therefore exclude the Should Return Answer category from the final benchmark, yielding the 245 Should Re-search and 209 Should Visit Page samples used throughout this paper.

\subsection{Ruling Out Surface-Feature Confounds via Paired Flip Analysis}
\label{appendix:flip_analysis}

Because Agentic Mode and Observer Mode share exactly the same task text and search results, differing only in whether the results are framed as the outcome of the model's own prior action, our paired design already controls for the main effects of task content and search-result content. However, it does not by itself rule out an interaction between certain sample characteristics and the presentation mode. We therefore analyze \emph{pattern flips}: samples in the Should Re-search subset that a given model answers correctly in one mode but incorrectly in the other, evaluated across the six general-purpose models studied in Figure~\ref{fig:ibis_results}.

Empirically, flips are highly consistent in direction: 91.9\% of flips point toward the inertia direction, i.e., the model correctly re-searches in Observer Mode but stays on the current search path in Agentic Mode, and this directional pattern holds consistently across models. To test whether this pattern could instead be explained by properties of the search results themselves, we extract 18 shallow surface features from each sample's search results (e.g., number of results, snippet/title length, presence of date markers, query--snippet lexical overlap) and use them to predict (i) the direction of a flip and (ii) whether a given sample flips at all. Neither is predictable from these features: AUC $=0.498$ ($p=0.56$) for flip direction and AUC $=0.492$ ($p=0.64$) for flip occurrence, and no single feature's correlation with the per-item Agentic-vs-Observer gap survives multiple-comparison correction. These results indicate that the measured inertia-bias gap cannot be explained by the 18 shallow distributional features we measured, supporting our claim that it reflects action-history ownership rather than a search-result distribution artifact.

\section{Additional Evidence for the Necessity of Isolation in Search}
\begin{table}[t]
    \centering
    \caption{The effect of the context-isolated filter module on search efficiency. The table reports the average number of searches conducted by SearchAgent before and after the application of filtering. f/\O\ denotes that no dedicated filter module is used, f/\(\checkmark\) denotes that the standard filter module is used, and f/\(\square\) denotes that the context-isolated filter module is used.}
    \label{tab:filter_eval_main}
    \resizebox{\columnwidth}{!}{
      \begin{tabular}{@{}lcc@{}}
        \toprule
        Method & Before Filter & After Filter\,$\downarrow$ \\
        \midrule
        NIS-Agent\textsuperscript{f/\O} & 9.8 & 9.8\phantom{000} \\
        NIS-Agent\textsuperscript{f/\(\checkmark\)} & 9.6 & 8.9\uagdown{9.18} \\
        NIS-Agent\textsuperscript{f/\(\square\)} & 9.4 & \textbf{7.6}\uagdown{22.45} \\
        \bottomrule
      \end{tabular}}
\end{table}

We specifically investigate why a dedicated context-isolated mechanism is necessary, rather than simply integrating a filtering step within the agent's standard workflow. As established in Section~\ref{sec:ibis}, agents suffer from inertia bias—a tendency to rationalize their own prior actions. Consequently, when an agent is asked to self-evaluate search results within its own execution context, it lacks the objectivity to reject irrelevant information. Even when explicitly instructed to aggressively filter out non-essential content, the standard integrated approach yields suboptimal results compared to our isolated mechanism.

To this end, we conducted an additional set of contrastive experiments. For the filter module in SearchAgent, we adopt the GAIA subset mentioned in Section~\ref{main_results}, which comprises 50 queries, and conducted a simple comparative experiment based on NIS-Agent. With GPT-4o as the base model, the results in Table~\ref{tab:filter_eval_main} show that our module effectively filters out irrelevant webpages.

\begin{figure*}[t!]
\begin{center}
\includegraphics[width=\textwidth]{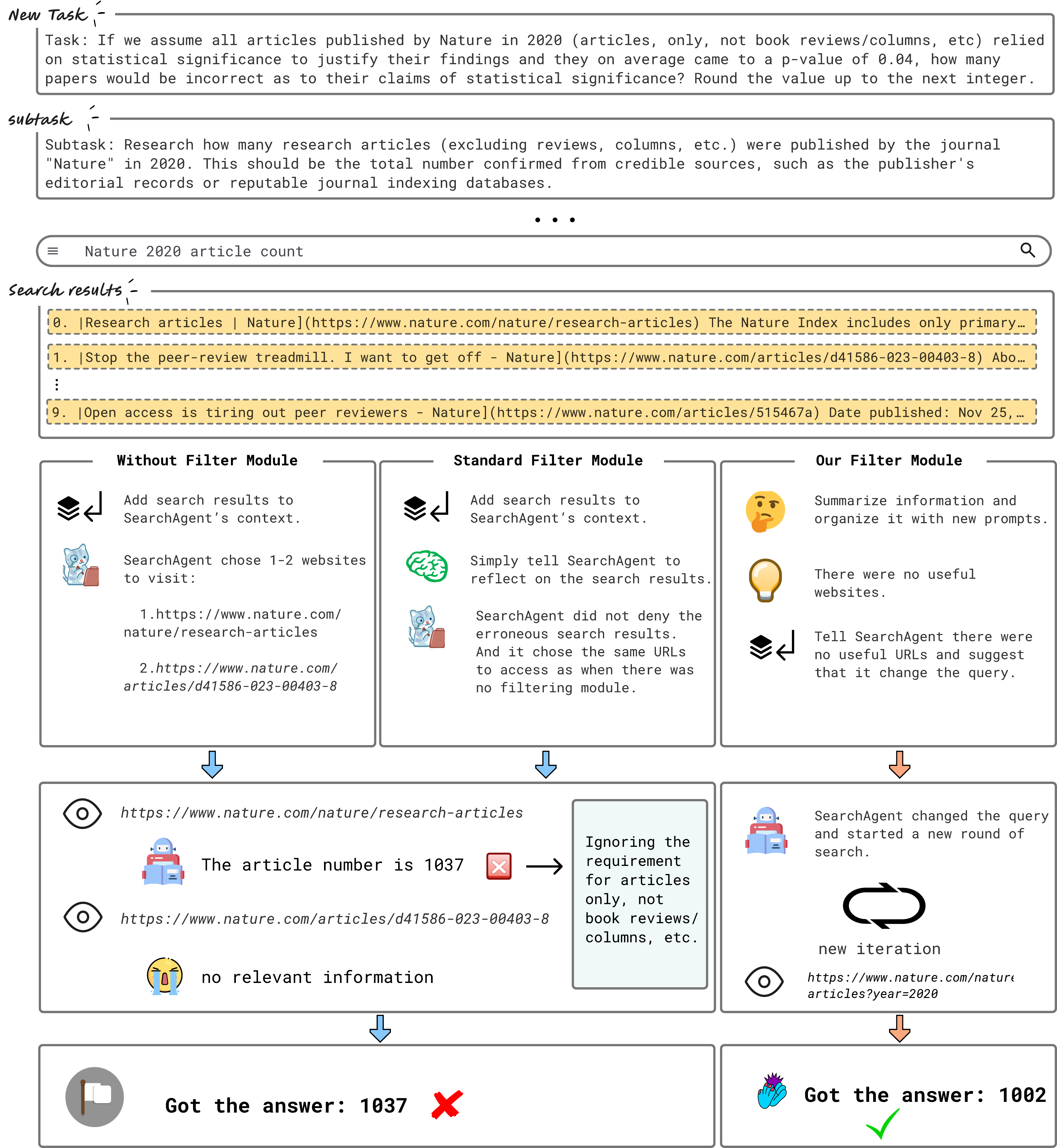}
\end{center}
\caption{A comparison of search filter modules and their effect on answer accuracy. After obtaining the search results, the SearchAgent may process them in different ways. Without a filter module, the LLM simply selects a few seemingly most relevant webpages to visit. With a standard filter module, however, the selected pages are often still those that appear relevant but are in fact irrelevant, which is almost the same as the first case. In contrast, our filter module isolates the memory of the SearchAgent and retains only the most relevant content for evaluation, enabling a more objective judgment of relevance. It can also perform adaptive rewriting and re-iteration in time, thereby avoiding noise and making it easier to reach the correct answer.}
\label{contrast}
\end{figure*}

\subsection{Details of IBIS Experiments}
\label{Apendix:IBIS_Experiments}

\begin{table*}[t]
    \centering
    \small
    \caption{Model performance comparison on IBIS. The results represent the accuracy of the next-step decision. For the \textit{Should Visit Page} task, calling the \texttt{visit\_page} tool is considered correct. Conversely, for the \textit{Should Re-search} task, calling the \texttt{web\_search} tool is considered correct (whereas calling \texttt{visit\_page} is counted as an error).}
    \label{tab:all_IBIS_model_performance}
    \resizebox{\textwidth}{!}{%
        \begin{tabular}{lcccccc}
            \toprule
            \multirow{2}{*}{\textbf{Model}} & \multicolumn{3}{c}{\textbf{Task: Should Visit Page}} & \multicolumn{3}{c}{\textbf{Task: Should Re-search}} \\
            \cmidrule(lr){2-4} \cmidrule(lr){5-7}
             & \textbf{Agentic Mode} & \textbf{Observer Mode} & \textbf{Direct Mode} & \textbf{Agentic Mode} & \textbf{Observer Mode} & \textbf{Direct Mode} \\
            \midrule
            \textbf{Claude-3.7-Sonnet}  & 97.61  & 89.95 & 97.61 & 66.12 & 87.35 & 75.51 \\
            \textbf{deepseek-v4-pro}  & 95.22  & 91.87 & 93.30 & 67.76 & 81.63 & 79.59 \\
            \textbf{Gemini-2.5-Pro}     & 98.09  & 96.17 & 98.56 & 31.43 & 62.44 & 46.12 \\
            \textbf{GPT-4.1-2025-04-14} & 97.61  & 85.65 & 96.17 & 20.41 & 59.18 & 86.93 \\
            \textbf{GPT-4o-2024-11-20}  & 100.00 & 91.86 & 91.39 & 4.08  & 25.31 & 71.83 \\
            \textbf{Qwen3-max}          & 97.61  & 96.65 & 99.52 & 26.12 & 41.22 & 50.61 \\
            \textbf{NIS-8B}             &  97.61   & 91.87  &  92.34 &  66.94 & 68.57 & 73.46  \\
            \bottomrule
        \end{tabular}%
    }
\end{table*}

We evaluate the 5 models mentioned in Section~\ref{sec:exp_setup} and our trained model NIS-8B on the \textit{Should Visit Page} subset. Furthermore, we employ our Context-Isolated Filter Module (Direct Mode) to independently assess the performance of different LLMs on the IBIS benchmark. The results are presented in Table~\ref{tab:all_IBIS_model_performance}. From the table, we observe the following:
\begin{itemize}
    \item \textbf{Divergent Model Tendencies.} Models exhibit distinct behavioral biases regarding the next step. For instance, Claude-3.7-Sonnet maintains a more cautious stance, showing a higher propensity to call the \texttt{web\_search} tool to refine results. In contrast, GPT-4o demonstrates extreme confidence in its initial queries, heavily favoring the \texttt{visit\_page} tool, which leads to a collapse in performance on the \textit{Should Re-search} task (only 4.08\% accuracy in Agentic Mode).

    \item \textbf{The ``Confidence'' from Inertia Bias.} Interestingly, on the \textit{Should Visit Page} subset, models in Agentic Mode consistently outperform those in Observer Mode. This does not imply superior reasoning in Agentic Mode; rather, it confirms the presence of inertia bias. Knowing its own action history, the model feels a ``commitment'' to its generated query and is statistically more inclined to click a page (\texttt{visit\_page}). This tendency artificially boosts accuracy when the page \textit{should} be visited, but causes significant failures when the results are irrelevant (as seen in the \textit{Should Re-search} subset). This contrast proves that the high performance in Agentic Mode on this subset is partly driven by bias rather than objective judgment.

    \item \textbf{Validation of the Direct Mode Design.} The Direct Mode is designed to evaluate relevance judgment by isolating the current query and candidate webpages from the model's action history. On the \textit{Should Visit Page} subset, Direct Mode achieves accuracy comparable to Agentic Mode and significantly higher than Observer Mode. This indicates that Direct Mode successfully avoids the performance degradation seen in Observer Mode, which stems from detaching the observation from the model's own context. We argue that avoiding ``search noise'' (unnecessary browsing) must not come at the cost of missing relevant information. Therefore, a pure Observer-style design—which lowers accuracy on valid pages—is unacceptable. In contrast, Direct Mode strikes an optimal balance: it maintains high recall on potential answers (matching the ``confidence'' of Agentic Mode) while significantly reducing the intake of irrelevant webpages compared to the biased Agentic Mode.
\end{itemize}

\section{Details of NIS-Agent}

\subsection{Workflow of Context-Isolated Filter Module}

Figure~\ref{contrast} illustrates how NIS-Agent mitigates the inertia bias during the search phase.

\begin{figure}[t]
  \centering
  \includegraphics[width=\columnwidth]{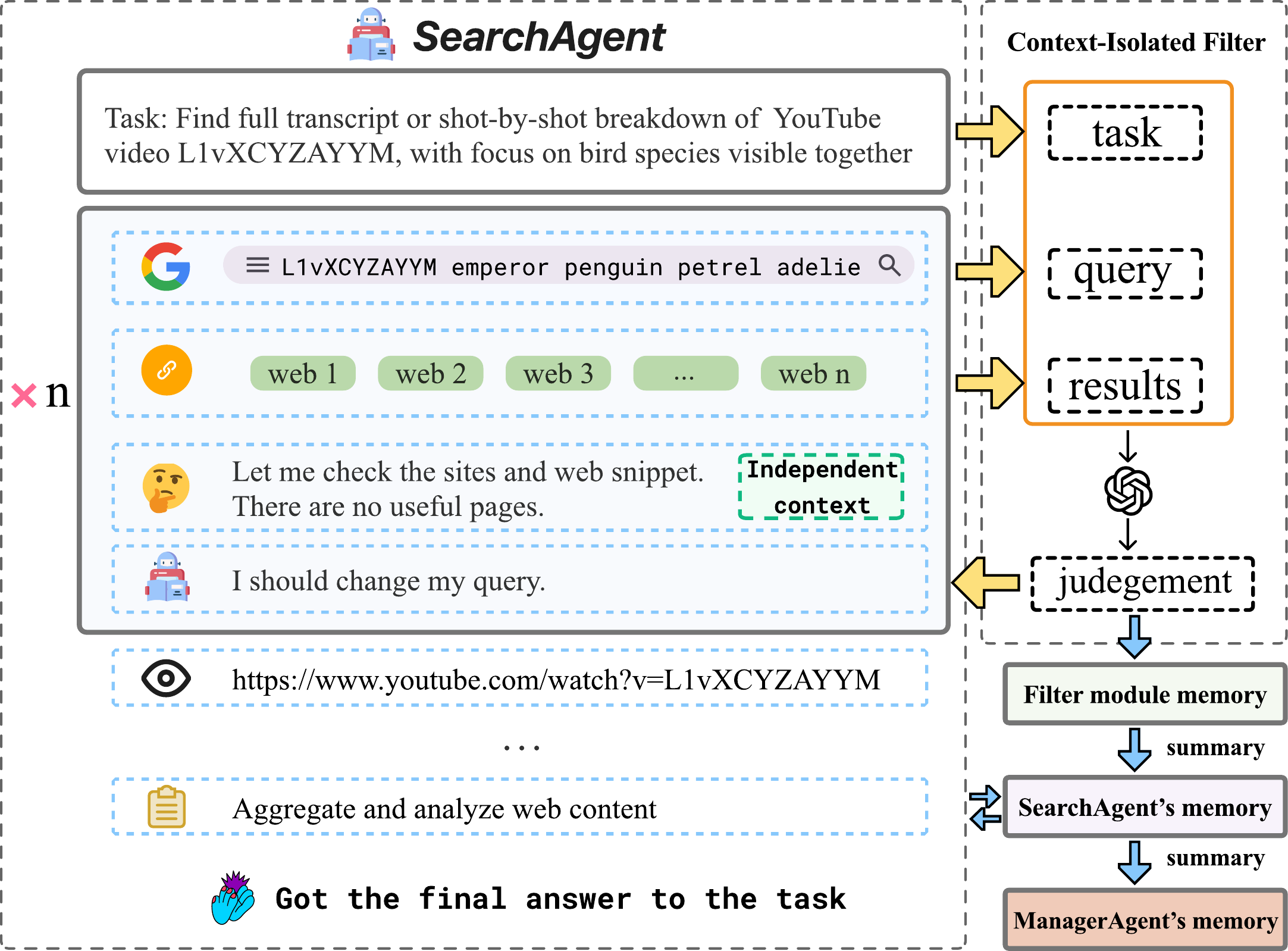}
  \caption{Workflow of the context-isolated filter module. It extracts the task, query, and results from the SearchAgent’s memory and evaluates their relevance.}
  \label{fig:web_filter_main}
\end{figure}

Figure~\ref{fig:web_filter_main} illustrates the workflow of the context-isolated filter module.

\subsection{Web browser toolkit}
\label{appendix:web_toolkit}

\begin{table*}[t]
\footnotesize
\centering
\caption{Web browser toolkit}
\label{lab:web_browser_toolkit}
\begin{tabularx}{\textwidth}{lXl}
\toprule
\textbf{Tool Name} & \textbf{Description} & \textbf{Input Parameters} \\
\midrule
web\_search & Perform Google searches and retrieve search results & \textit{query}, \textit{filter\_year (optional)} \\
fetch\_html & Read and analyze the content of an HTML page & \textit{url}, \textit{query} \\
fetch\_pdf & Read and analyze the content of a PDF page & \textit{url}, \textit{query} \\
visit\_page & Returns transcript if the link is YouTube, otherwise downloads the web resource & \textit{url} \\
find\_archived\_url & Finds historical content through web archive services & \textit{url}, \textit{date} \\
inspect\_file\_as\_text & Reads the file as text and answers questions & \textit{file\_path}, \textit{question (optional)} \\
\bottomrule
\end{tabularx}
\end{table*}

Table~\ref{lab:web_browser_toolkit} presents all search-related tools integrated into the SearchAgent.

\subsection{Implementation Details}
\label{sec:implementation details}
On the GAIA and WebWalkerQA benchmarks, we set the temperature of all models to 0.0. For non-reasoning models, we limit the output token size to 4096. We implement the isolation-based stepwise validation module as a ValidationAgent, while conceptually it remains a module within the overall framework.

In terms of tool integration, we have authorized multiple Python standard libraries and third-party libraries for model code execution. The model can dynamically load and invoke the following libraries, ensuring that NIS-Agent is capable of handling complex input formats, including text, images, tables, and web data.

\begin{center}
	\begin{minipage}{1\linewidth}
		\begin{multicols}{3}         % <-- 添加这一行，设置为 3 栏    
			\setlength\itemsep{2pt}     
			\begin{itemize}[leftmargin=*]
				\item requests
				\item zipfile
				\item os
				\item pandas
				\item numpy
				\item sympy
				\item json
				\item bs4
				\item pubchempy
				\item xml
				\item yahoo\_finance  
				\item Bio
				\item sklearn
				\item scipy
				\item pydub
				\item io
				\item PIL
				\item chess
				\item PyPDF2
				\item pptx
				\item torch
				\item datetime
				\item fractions
				\item csv
			\end{itemize}
		\end{multicols}        % <-- 添加这一行
	\end{minipage}
\end{center}

\vspace{1em}

For both benchmarks, we report only the \textit{Pass@1} results, and all evaluations strictly follow the experimental protocols defined by GAIA and WebWalkerQA. To ensure objectivity, we restrict our comparison of baselines to single-attempt performance metrics. This includes explicitly reported \textit{Pass@1} results and average accuracy scores from multiple experimental runs.

\subsection{Details of Deep Research Experiments}
During the evaluation of smolagents DR, we modified the original smolagents DR code and integrated (or ported) some of NIS-Agent's tools into smolagents DR to ensure a fair comparison. For example, since the original smolagents tool for viewing XLSX files could not recognize colors, we redeveloped a tool capable of identifying both colors and background graphics, and implemented it concurrently in both NIS-Agent and smolagents.

\subsubsection{Sources of Deep Research Results}
All experimental results reported in paper are taken from the corresponding official papers. The only exception is the WebWalkerQA result for MiroFlow: since the original MiroFlow paper does not report a WebWalkerQA score, we use the reproduced result reported by \citet{chen2026searchmorethinkless}.

\subsubsection{Subset Reliability on WebWalkerQA}

Due to the high cost of deep research tasks, evaluating on large benchmarks in full is prohibitive, and nearly all prior methods report results on a randomly sampled subset (e.g., WebExplorer, BrowseMaster). Our use of 200 randomly sampled queries is consistent with this standard practice. To verify that our subset reliably reflects full-benchmark performance, we additionally ran NIS-Agent with Claude-3.7-Sonnet on the complete WebWalkerQA dataset (680 queries in total). The full-dataset accuracy is 69.26\%, which is largely consistent with the 68.50\% obtained on our 200-query subset. This confirms that random sampling of 200 queries provides a reliable estimate of performance on the full benchmark.

\subsubsection{Run time}

\begin{figure}[t]
    \centering
    \includegraphics[width=\columnwidth]{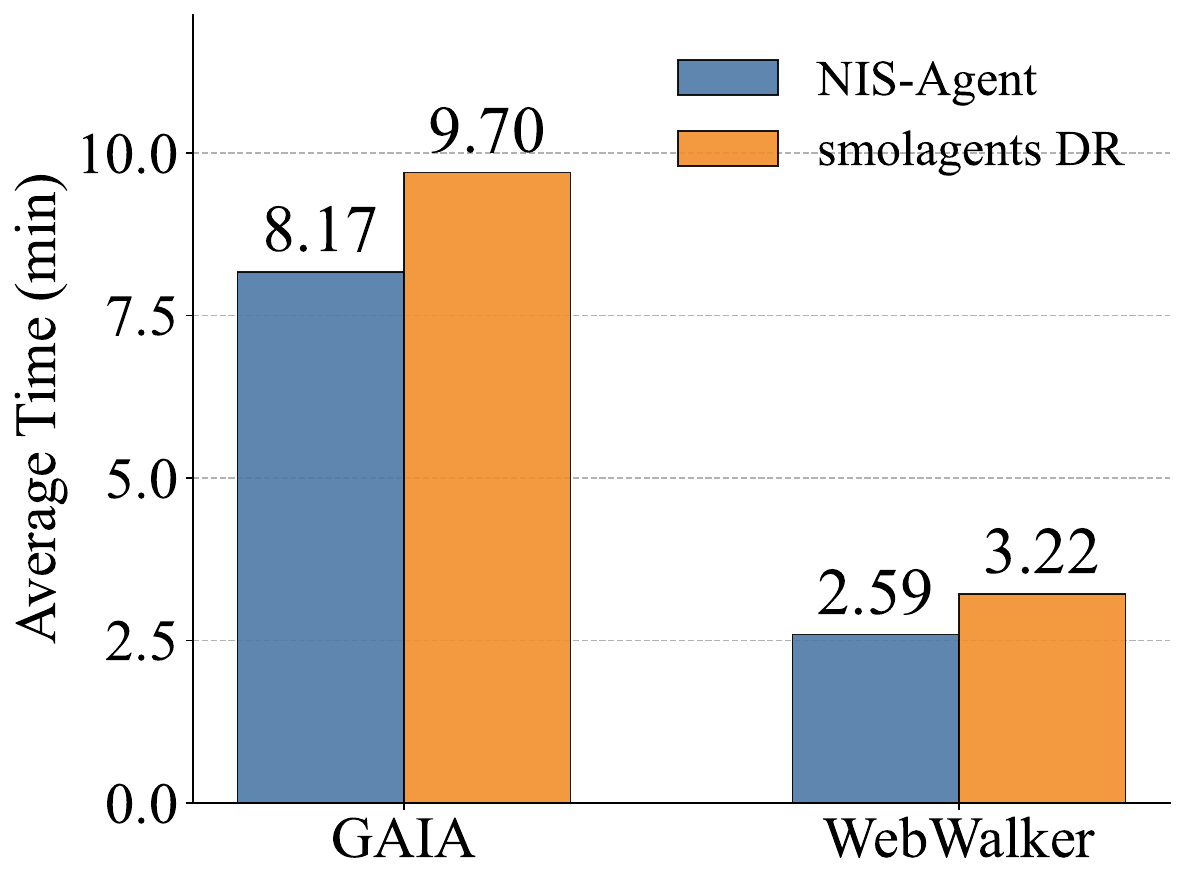}
    \caption{Average runtime per task.}
    \label{fig:run_time}
\end{figure}

As shown in Figure~\ref{fig:run_time}, the average runtime of NIS-Agent is lower than that of smolagents DR. Furthermore, with the increase in task difficulty, the runtime savings achieved by NIS-Agent become more pronounced.

\subsubsection{Qualitative Failure Case Analysis}
\label{appendix:failure_analysis}

We manually inspect the trajectories of remaining failures on GAIA to understand where NIS-Agent still falls short. Failures attributable to context isolation itself are rare. The most common pattern occurs in the context-isolated filter module: a retrieved abstract appears only weakly relevant to the question, while the key evidence actually lies in the body of the page, and the filter mistakenly discards the page as irrelevant. For example, on a question that asks for a specific figure disclosed deep in a company's annual report, the search snippet only mentions the report's title and publication date, with no visible numeric content, so the filter judges the page as unrelated and skips it. Because NIS-Agent proceeds iteratively, a single such misjudgment does not always cause the task to fail outright; the agent can often recover through a follow-up search that surfaces the same evidence from a different page. For the isolation-based stepwise validation module, we do not observe a comparably recurring error pattern, since the ManagerAgent still sees the full context when acting on the validation module's suggestions and can override a suggestion it judges unreasonable.

Beyond these isolation-specific cases, the large majority of remaining failures fall into four categories that are largely independent of context isolation. First, some tasks demand reasoning that exceeds the backbone model's own ability, for instance multi-step arithmetic over figures scattered across several sources, so the agent retrieves the correct evidence but still derives the wrong final answer. Second, some tasks require chaining many search hops before the key evidence appears, and the agent exhausts its step budget before reaching it. Third, the target page for some tasks is very long, and the answer is buried in a location that keyword-based localization on the text browser fails to surface. Fourth, some target pages depend on JavaScript rendering or other dynamic content that our text-based browser tool cannot faithfully reproduce, so the relevant content is never made available to the agent in the first place.

\section{Details of NIS-8B Training}
\label{sec:nis8b_training}

\paragraph{Data.}
Both SFT and RL data follow the IBIS construction protocol: each prompt presents a multi-hop question, the agent's first \texttt{web\_search} action, and a set of retrieved abstracts designed to require a non-trivial next decision. The SFT corpus contains 4{,}293 examples; the RL corpus contains 9{,}964 examples. Both are disjoint from our evaluation benchmarks.

For SFT, each prompt is paired with a gold action produced by a strong external teacher operating in Observer Mode, providing direct behavioral supervision on the inertia-bias decision point.

For RL, gold actions are not used as fixed supervised targets. Instead, each prompt is associated with a structured \textbf{reference rubric} generated offline by a strong evaluator (Claude Sonnet 4), and each rollout's chosen action is graded online against this rubric during training.

\paragraph{Two-stage judge.}
A naive design would invoke a strong Observer-Mode LLM for every rollout, asking it to re-read the search results and judge the action in one pass. At GRPO group size $K{=}32$, this scales linearly with the number of rollouts and dominates the training cost. We instead decouple the judge into two stages:

\textit{Stage~1 (offline).}\quad
Before training, we call a strong evaluator (Claude\,Sonnet\,4 in our setup) \emph{once per prompt} with the same isolation-conditioned instruction used by the Observer-Mode evaluator at inference. Rather than returning a verdict, the evaluator emits a structured \emph{reference rubric} containing
(i)~the correct next action category (\textit{open\_result}, \textit{search\_again}, or \textit{final\_answer}),
(ii)~the acceptable evidence URLs or acceptable reframed queries that would qualify as correct,
(iii)~a list of disqualifying failure modes,
and (iv)~a one-sentence grading instruction. This rubric is stored alongside the prompt as fixed metadata.

\textit{Stage~2 (online).}\quad
At training time, each rollout's chosen action is scored by a cheap model (qwen3-max) that receives only the research question, the agent's prior query, the reference rubric, and the chosen action. The cheap model never sees the raw search results; its task is reduced from open-ended Observer-Mode reasoning to a short rule-following pass over the rubric.

We grade on \emph{action class} as the primary signal. The cheap model maps the agent's tool invocation to one of three classes---\textit{open\_result} (\texttt{jina\_fetch\_html} / \texttt{jina\_fetch\_pdf}), \textit{search\_again} (\texttt{web\_search}), or \textit{final\_answer} (\texttt{<final\_answer>})---and compares this class to the rubric's prescribed next-action category. Tool-class mismatch (or a malformed action that does not parse into any class) is the only path to $R_{\text{action}}=0$. When the class matches, a soft secondary check on the parameter distinguishes a clean choice from a degenerate one: an empty or obviously off-topic URL, a re-search query that merely echoes the prior query, or a final answer that is empty or unrelated to the grading instruction is downgraded to $R_{\text{action}}=0.7$ rather than scored as incorrect. The judge therefore returns a \emph{ternary} verdict at temperature 0,
$$R_{\text{action}} \in \{0,\, 0.7,\, 1\},$$
which we found necessary because the rubric's $\textit{acceptable\_evidence}$ and $\textit{acceptable\_new\_queries}$ lists are inevitably non-exhaustive; a hard binary cutoff at the parameter level discards partial-credit signal that the policy can still learn from. The rubric's $\textit{bad\_actions}$ list is treated as a hard veto and forces an \textit{incorrect} verdict regardless of class.

Because the rubric is generated once per prompt and reused across $K$ rollouts and across epochs, the expensive call is a fixed cost amortized over the entire training run, while the per-rollout cost is held to a single short cheap-model call.

\section{Details of IBIS Evaluation Modes}
\label{appendix:ibis_prompts}

In this section, we introduce the implementation details of the two modes.

\paragraph{Observer Mode.}
In this configuration, the model is positioned as a neutral evaluator. The search results are embedded directly into the user's input as external context, with no conversational history indicating that the model initiated the search.

\vspace{1em}

\begin{lstlisting}
messages = [
    {
        "role": "system", 
        "content": SYSTEM_PROMPT,
        "cache_control": {"type": "ephemeral"}
    },
    {
        "role": "user",
        "content": [
            {
                "type": "text", 
                "text": f"Task: {task_theme}"
            },
            {
                "type": "text", 
                "text": f"Here are some search results that may help you with this task:\n```\n{search_results}\n```"
            },
            {
                "type": "text",
                "text": "<system-reminder>You should first consider whether these webpages are helpful for completing the task based on the summaries. If they are not helpful, you need to search with a different query. Your response should always be a tool call.</system-reminder>"
            }
        ]
    }
]
\end{lstlisting}

\paragraph{Agentic Mode.}
In this configuration, we construct a synthetic conversation history to trigger the inertia bias. We explicitly inject an \texttt{assistant} message containing the tool call and a subsequent \texttt{user} message containing the observation. This structure forces the model to evaluate the consequences of its "own" prior action:

\vspace{1em}

\begin{lstlisting}
messages = [
    {
        "role": "system", 
        "content": SYSTEM_PROMPT,
        "cache_control": {"type": "ephemeral"}
    },
    {
        "role": "user", 
        "content": f"Task: {task_theme}"
    },
    {
        "role": "assistant",
        "content": [
            {
                "type": "text",
                "text": f"Calling tools:\n"
                        f"[{{'id': '{tool_call_id}', "
                        f"'type': 'function', "
                        f"'function': {{'name': 'web_search', "
                        f"'arguments': {{'query': '{query}'}}}}}}]"
            }
        ]
    },
    {
        "role": "user",
        "content": [
            {
                "type": "text",
                "text": f"Call id: {tool_call_id}\nObservation:\n```\n{search_results}\n```"
            },
            {
                "type": "text",
                "text": "<system-reminder>You should first consider whether these webpages are helpful for completing the task based on the summaries. If they are not helpful, you need to adjust the query and search again. Your response should always be a tool call.</system-reminder>"
            }
        ]
    }
]
\end{lstlisting}

Note that an identical \texttt{<system-reminder>} is appended in both modes to ensure the model is explicitly aware of its option to reject the results.

\end{document}